\documentclass[sigconf]{acmart}
\AtBeginDocument{%
  }

\usepackage{multirow}
\usepackage{graphicx}
\usepackage{float}
\usepackage{subcaption} 
\usepackage{algorithm}
\usepackage{algorithmic}
\usepackage{amsmath}
\usepackage{xcolor}

\setcopyright{acmlicensed}
\copyrightyear{2026}
\acmYear{2026}
\acmDOI{XXXXXXX.XXXXXXX}
\acmConference[Conference acronym 'XX]{Make sure to enter the correct
  conference title from your rights confirmation email}{June 03--05,
  2018}{Woodstock, NY}
\acmISBN{978-1-4503-XXXX-X/2018/06}

\begin{document}

\title{ExpressMind: A Multimodal Pretrained Large Language Model for Expressway Operation}

\author{Zihe Wang}
\orcid{1234-5678-9012}
\affiliation{%
  \institution{Beihang University}
  \city{Beijing}
  %\state{Beijing}
  \country{China}}
\email{by2313310@buaa.edu.cn}

\author{Yihuan Wang}
\affiliation{%
  \institution{Beihang University}
  \city{Beijing}
  \country{China}}
\email{yhuanwang@buaa.edu.cn}

\author{Haiyang Yu}
\affiliation{%
  \institution{Beihang University}
  \city{Beijing}
  \country{China}}
\email{hyyu@buaa.edu.cn}

\author{Zhiyong Cui}
\authornote{Corresponding author}
\affiliation{%
  \institution{Beihang University}
  \city{Beijing}
  \country{China}}
\email{zhiyongc@buaa.edu.cn}

\author{Xiaojian Liao}
\affiliation{%
  \institution{Beihang University}
  \city{Beijing}
  \country{China}}
\email{liaoxj@buaa.edu.cn}

\author{Chengcheng Wang}
\affiliation{%
  \institution{Shandong Hi-speed Group Co., Ltd}
  \city{Jinan}
  \country{China}}
\email{wangchengcheng@sdhsg.com}

\author{Yonglin Tian}
\affiliation{%
  \institution{Institute of automation, Chinese Academy of Sciences}
  \city{Beijing}
  \country{China}}
\email{tyldyx@mail.ustc.edu.cn}

\author{Yongxin Tong}
\affiliation{%
  \institution{Beihang University}
  \city{Beijing}
  \country{China}}
\email{yxtong@buaa.edu.cn}

\renewcommand{\shortauthors}{Wang et al.}

\begin{abstract}
 The current expressway operation relies on rule-based and isolated models, which limits the ability to jointly analyze knowledge across different systems. Meanwhile, Large Language Models (LLMs) are increasingly applied in intelligent transportation, advancing traffic models from algorithmic to cognitive intelligence. However, general LLMs are unable to effectively understand the regulations and causal relationships of events in unconventional scenarios in the expressway field. Therefore, this paper constructs a pre-trained multimodal large language model (MLLM) for expressways, ExpressMind, which serves as the cognitive core for intelligent expressway operations. This paper constructs the industry’s first full-stack expressway dataset, encompassing traffic knowledge texts, emergency reasoning chains, and annotated video events to overcome data scarcity. This paper proposes a dual-layer LLM pre-training paradigm based on self-supervised training and unsupervised learning. Additionally, this study introduces a Graph-Augmented RAG framework to dynamically index the expressway knowledge base. To enhance reasoning for expressway incident response strategies, we develop a RL-aligned Chain-of-Thought (RL-CoT) mechanism that enforces consistency between model reasoning and expert problem-solving heuristics for incident handling. Finally, ExpressMind integrates a cross-modal encoder to align the dynamic feature sequences under the visual and textual channels, enabling it to understand traffic scenes in both video and image modalities. Extensive experiments on our newly released multi-modal expressway benchmark demonstrate that ExpressMind comprehensively outperforms existing baselines in event detection, safety response generation, and complex traffic analysis. The code and data are available at: https://wanderhee.github.io/ExpressMind/.
\end{abstract}

\begin{CCSXML}
<ccs2012>
 <concept>
  <concept_id>00000000.0000000.0000000</concept_id>
  <concept_desc>Do Not Use This Code, Generate the Correct Terms for Your Paper</concept_desc>
  <concept_significance>500</concept_significance>
 </concept>
 <concept>
  <concept_id>00000000.00000000.00000000</concept_id>
  <concept_desc>Do Not Use This Code, Generate the Correct Terms for Your Paper</concept_desc>
  <concept_significance>300</concept_significance>
 </concept>
 <concept>
  <concept_id>00000000.00000000.00000000</concept_id>
  <concept_desc>Do Not Use This Code, Generate the Correct Terms for Your Paper</concept_desc>
  <concept_significance>100</concept_significance>
 </concept>
 <concept>
  <concept_id>00000000.00000000.00000000</concept_id>
  <concept_desc>Do Not Use This Code, Generate the Correct Terms for Your Paper</concept_desc>
  <concept_significance>100</concept_significance>
 </concept>
</ccs2012>
\end{CCSXML}

\ccsdesc[500]{Computing methodologies~Artificial intelligence}

\keywords{Large Language Models, Intelligent Expressway Operations, Pre-training Paradigm, Chain-of-Thought, Multimodal Understanding}

\maketitle

\section{Introduction}

\begin{figure}[htbp]
    \centering
    % width 可以设置为 \linewidth (单栏宽度) 或 \textwidth (页面宽度)
    % 如果觉得图太大，可以乘以系数，例如 0.9\linewidth
    \includegraphics[width=1.0\linewidth]{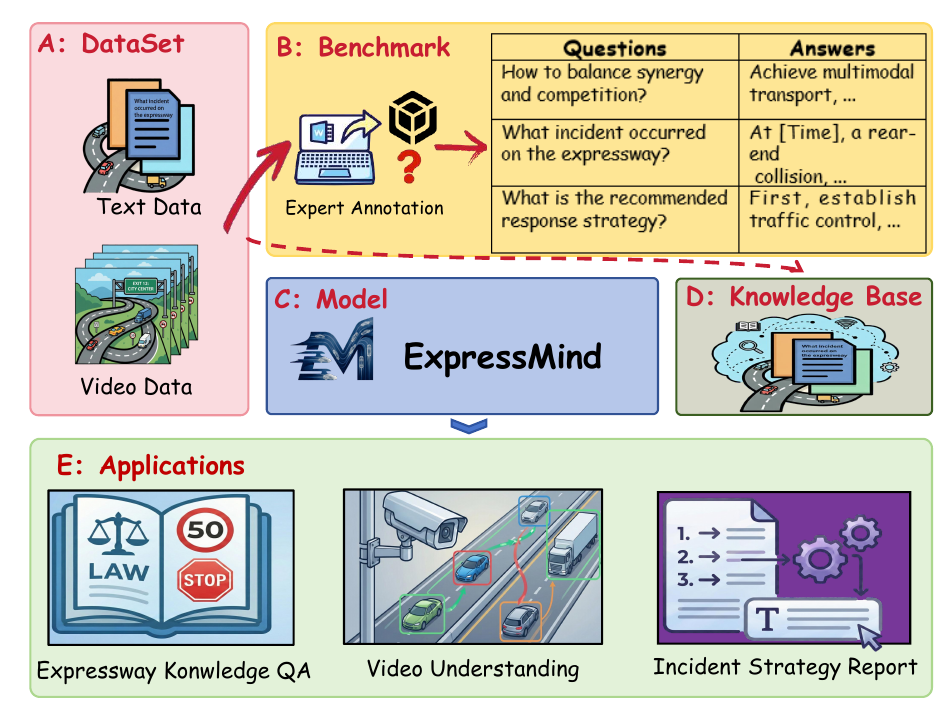}
    \caption{Overview of ExpressMind.}
    \label{fig:contributions}
\end{figure}

With the continuous advancement of intelligent transportation systems (ITS), expressway operation is evolving from the reactive and rule-based paradigm towards intelligent agents endowed with deep cognitive reasoning capabilities. Breakthroughs in artificial intelligence \cite{zhang2025htfllib}, particularly the emergence of Large Language Models (LLMs) with exceptional computational and reasoning abilities \cite{zhang2025fedmetro}, are profoundly shaping the intelligent transformation across ITS industries. However, the expressway sector still lacks a domain-specific LLM which is capable of deeply adapting to complex expressway operational needs. At the current stage, advancing foundation models' core capabilities such as expressway knowledge integration, multimodal scene understanding, and autonomous incident reasoning remains a critical bottleneck, necessitating a fundamental methodological shift.

Obviously, most open-source LLMs \cite{ma2025overcoming} lack a deep understanding of publicly inaccessible specialized knowledge in the expressway domain, such as technical standards and regulations. Dynamic information and professional terminology cannot be promptly fed into LLMs and, thus, their reasoning and decision-making processes struggle to guarantee safety and efficiency requirements. Furthermore, existing methods exhibit shortcomings in key visual feature extraction and traffic-related reasoning within multimodal scenarios. Therefore, current approaches fail to meet the dynamic and precise operation requirements of expressway, lacking an intelligent central hub capable of multi-task collaborative cognition and deep industry understanding.

However, constructing an expressway domain-specific multimodal large model as such intelligent central hub faces considerable difficulties and unique challenges. The highly heterogeneous and complex multimodal data in this field is the first hurdle. The unstructured data such as real-time monitoring videos and semi-structured data including incident records, making it extremely difficult to achieve effective alignment and fusion between different modalities. Additionally, the expressway domain has strict requirements on safety and accuracy. The desired foundation model needs to accurately grasp professional knowledge such as traffic engineering principles and emergency disposal specifications, which brings great challenges to the in-depth integration of domain knowledge and the design of multimodal architecture. More importantly, the scarcity of high-quality labeled multimodal data in the expressway field, coupled with the privacy and security constraints, further increases the difficulty of model training and optimization, becoming a key obstacle to building a high-performance domain-specific multimodal large model.

% Developing a domain LLM for expressway operation presents several challenges. There is a scarcity of high-quality annotated domain-specific data. The operational environment is highly dynamic and filled with numerous long-tail scenarios, which complicates reliable causal reasoning and compliant decision-making. Existing LLMs have a limited ability to effectively align spatiotemporal patterns with traffic rules and domain knowledge. Therefore, developing a dedicated LLM for the expressway domain is imperative.

To address these challenges, this paper introduces ExpressMind, a domain multimodal LLM for expressway operation. We construct the first full-stack expressway dataset and propose a two-stage pre-training paradigm for the internalization of expressway-domain knowledge. This study also develops a Reinforcement Learning (RL)-based Chain-of-Thought (CoT) alignment mechanism to strengthen domain reasoning. Furthermore, a visual-enhanced cross-modal encoder is incorporated and a graph-based retrieval-augmented generation (RAG) is proposed to enhance the extraction of key traffic scene characteristics and dynamic knowledge. 
% To enable precise retrieval of accurate information and dynamic knowledge enhancement, this study built an expressway-domain knowledge base and adopted a graph-based retrieval-augmented generation method. 
The integration of these modules as a whole build the foundation of the multimodal pretrained LLM to process multi-source expressway data and provide efficient expressway operation decision support.

The overview of ExpressMind is illustrated in Figure. 
\ref{fig:contributions}, and its five core contributions are summarized as follows: 

\begin{itemize}
    \item \textbf{Full-stack Expressway dataset}: This study constructs the first industry's full-stack expressway dataset spanning text cognition, logical reasoning, and visual perception, including three specialized subsets: traffic knowledge texts, emergency response reasoning, and event video scene understanding.

    \item \textbf{RL-aligned CoT Reasoning}: We design a RL-based expressway strategy alignment strategy in LLM training, which can significantly enhance the model's logical reasoning and self-correction capabilities.
    
    \item \textbf{Graph-Augmented Retrieval}: A graph RAG-based dynamic knowledge base is established for critical expressway information retrieval and indexing.
    
    \item \textbf{Multimodal Alignment mechanism}: A Visual-Prior Alignment mechanism is designed by enforcing alignment and reweighting of visual tokens to enhance the understanding of visual features.
    
    \item \textbf{Multi-modal Benchmark}: The multi-modal Benchmark for evaluating LLMs within the expressway domain is released, encompassing four evaluation subsets: basic knowledge comprehension, video incident detection, safety response generation, and traffic analysis reporting.
\end{itemize}

\section{Relatedwork}

\begin{figure*}[t] % 加 * 号表示跨双栏，[t] 表示优先置顶
  \centering % 图片居中
  \includegraphics[width=\textwidth]{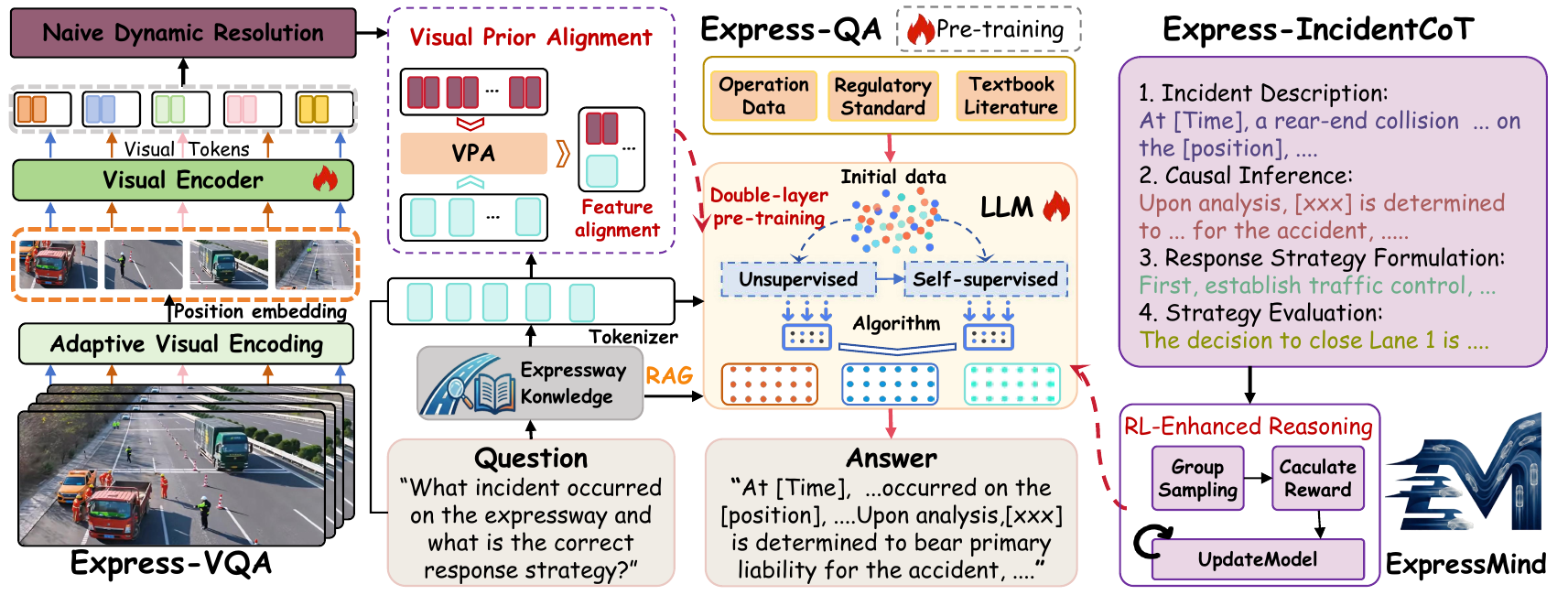}
  \caption{The Overall Framework of ExpressMind.}
  \Description{The architectural diagram of ExpressMind showing three key components: Adaptive Visual Encoding, RL-Enhanced CoT Reasoning, and Domain Knowledge Integration.}
  \label{fig:framework}
\end{figure*}

\noindent \textbf{Traffic related foundation model:} Current mainstream LLM architectures are primarily categorized into three paradigms: Encoder-only (e.g., BERT \cite{devlin-etal-2019-bert}), Encoder-Decoder (e.g., GLM \cite{glm2024chatglm}), and Decoder-only (e.g., GPT \cite{achiam2023gpt}, LLaMA \cite{dubey2024llama}). To adapt general-purpose models to vertical domains, methodologies such as Supervised Fine-Tuning (SFT) \cite{ouyang2022training} and Parameter-Efficient Fine-Tuning (PEFT) \cite{hu2022lora} have been introduced. Within the transportation sector, models such as TransGPT \cite{wang2024transgpt} and TrafficGPT \cite{zhang2024trafficgpt} advances traffic safety analysis via domain-adaptive training. Furthermore, UrbanGPT \cite{li2024urbangpt} integrates spatio-temporal encoders with LLMs through instruction tuning for general urban analysis. However, a domain LLM for expressway tasks has yet to emerge. 

\noindent \textbf{MLLMs:} CLIP \cite{radford2021learning} established the foundation of multimodal learning by aligning the representation spaces of images and texts via contrastive learning. To endow LLMs with visual comprehension capabilities, LLaVA \cite{liu2023visual} introduces a linear projection layer to map visual features into token embeddings processable by language models. BLIP-2 \cite{li2023blip} proposes the Q-former architecture to extract text-relevant features from frozen visual encoders.Following this research trajectory, the Qwen-VL series \cite{bai2025qwen2} is subsequently proposed to align visual and linguistic representations, In traffic scenarios, MLLMs have been applied to tasks such as accident analysis (TrafficLens \cite{arefeen2024trafficlens}, MoTIF \cite{wang2025motif}) and anomaly detection (Anomaly-OneVision \cite{xu2025towards}) by facilitating semantic understanding of surveillance footage.

\noindent \textbf{Reinforcement Learning for Reasoning:} Reinforcement Learning from Human Feedback \cite{ouyang2022training} has emerged as the prevailing paradigm for aligning LLMs with human intent. Algorithms such as DPO \cite{rafailov2023direct}, CPO \cite{xu2024contrastive}, and GRPO \cite{shao2024deepseekmath} leverage preference information inherent in CoT processes to further optimize reasoning trajectories and training efficiency. To mitigate reward hacking, DreamPRM \cite{cao2025dreamprm} introduces a domain re-weighting mechanism. Recently, integrating the semantic comprehension of LLMs with the decision-making capabilities of RL has moved to the forefront of transportation research. Traffic-R1 \cite{zou2025traffic} and LLMLight \cite{lai2025llmlight} employ RL to enhance the generalization of LLMs in signal control tasks. AgentsCoMerge \cite{hu2025agentscomerge} uses RL with ramp and density rewards for traffic optimization, Time-LLM \cite{jin2023time} transforms time‑series into text via RL, achieving SOTA in traffic forecasting. While RL has been applied in related traffic tasks, its use for enhancing reasoning in the expressway domain remains unexplored.

\section{Methodology}
This study proposes ExpressMind, a domain-specific MLLM tailored for expressway operation. The overall framework is illustrated in Figure.~\ref{fig:framework} and the key components are introduced as follows:

\subsection{Task-oriented Domain Data Profiling}
\begin{figure}[htbp]
    \centering
    % width 可以设置为 \linewidth (单栏宽度) 或 \textwidth (页面宽度)
    % 如果觉得图太大，可以乘以系数，例如 0.9\linewidth
    \includegraphics[width=1.0\linewidth]{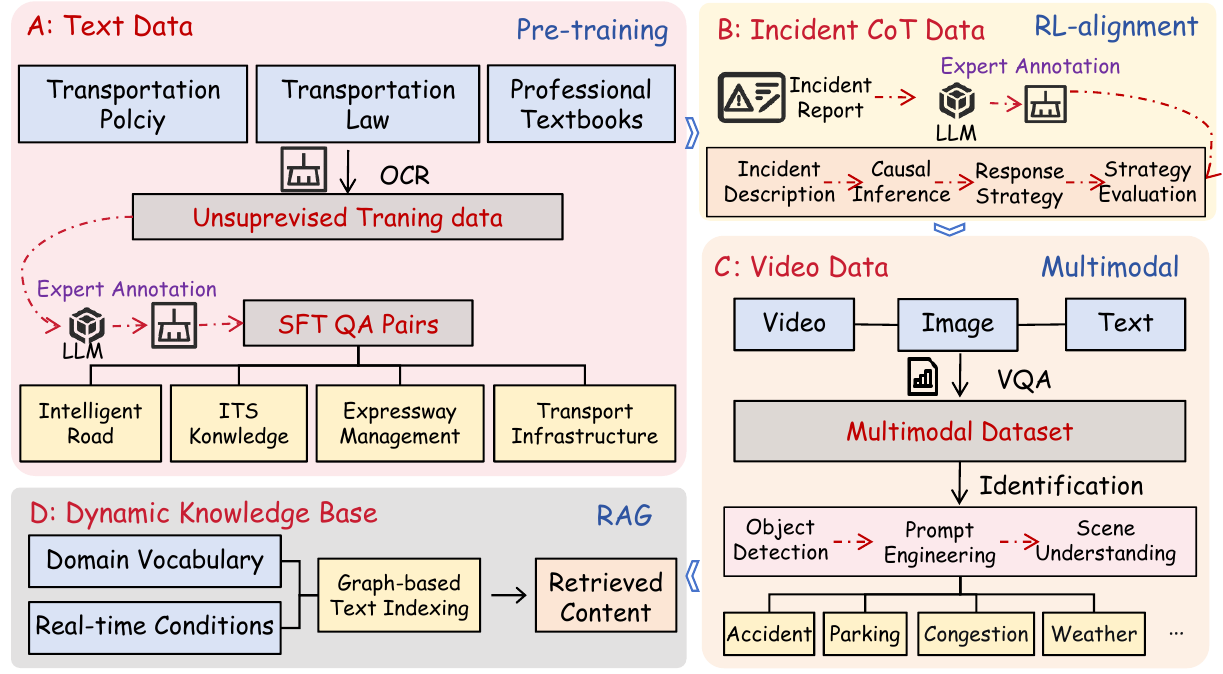}
    \caption{Task-oriented Domain Data Profiling.}
    \label{fig:data}
\end{figure}

To address the domain-specific tasks depicted above, which include Expressway Knowledge QA, Video Understanding, and Incident Strategy Report, this study collects four distinct types of data to support the complete training pipeline of ExpressMind, as illustrated in Figure~\ref{fig:data}.

\begin{itemize}
    \item \textbf{Textual Data}: To establish the model's fundamental domain understanding, textual data, including policy documents, expert knowledge, and SFT QA pairs, are used in the \textbf{\ref{sec:training} Pre-training Stage}.

    \item \textbf{Incident CoT Data}: To refine reasoning trajectories via reinforcement learning, incident CoT data comprises incident descriptions, causal reasoning, response strategies, and evaluations. It is employed during the \textbf{\ref{sec: RL} RL Alignment Stage}.

    \item \textbf{Dynamic Knowledge Base}: To ensure model responses remain aligned with the latest operational scenarios, it contains real-time traffic conditions, incident reports, and traffic flow data, providing \textbf{\ref{sec: RAG} real-time retrieval augmentation} across all training stages.

    \item \textbf{Multimodal Data}: To achieve video-language understanding, data such as accident images and congestion videos are introduced in the \textbf{\ref{sec: multimodal} Cross-modal Alignment Stage} to achieve video-language understanding.

\end{itemize}

\subsection{Training Paradigm of Pretrained LLM}
\label{sec:training}
To ensure the model acquires high-quality foundational knowledge for expressway scenarios, we constructed a dedicated dataset containing unlabeled text and self-supervised QA pairs, with all data undergoing rigorous deduplication and standardization. ExpressMind, built upon the Qwen foundation model, adopts a two-phase pre-training strategy: the first phase establishes fundamental scenario knowledge, and the second phase adapts the model to handle complex domain-specific tasks, respectively.

\paragraph{Stage 1: Unsupervised Training.} 
In this phase, model parameters $\theta$ are optimized by minimizing the negative log-likelihood loss. Given an input sequence $x = \{x_1, x_2, \dots, x_T\}$ derived from domain-specific corpora, the pre-training loss function $\mathcal{L}_{PT}$ is formulated as:

\begin{equation}
\mathcal{L}_{PT}(\theta) = - \sum_{t=1}^{T} \log P(x_t \mid x_{<t}; \theta)
\end{equation}
where $x_{<t}$ denotes the context sequence preceding time step $t$, and $P(x_t \mid x_{<t}; \theta)$ represents the conditional probability of the model predicting the next token given the current parameters.

\paragraph{Stage 2: Full-Parameter Supervised Fine-Tuning}
Following the acquisition of foundational domain knowledge, full-parameter SFT is conducted to align the model with specific tasks and instructions in the expressway transportation domain. During training, To ensure the model focuses on response generation, a masked loss strategy is employed by introducing a binary mask vector \(M\), where \(M_t = 0\) corresponds to instruction tokens and \(M_t = 1\) to response tokens. Consequently, the loss function for supervised fine-tuning, denoted as $\mathcal{L}_{SFT}$, is formulated as:
\begin{equation}
\mathcal{L}_{SFT}(\theta) = - \frac{1}{\sum_{t=1}^{T} M_t} \sum_{t=1}^{T} M_t \cdot \log P(x_t \mid x_{<t}; \theta)
\end{equation}

This two-stage training equips the model with an in-depth mastery of expressway domain knowledge, providing a basis for the following alignment and reasoning tasks.

\subsection{RL for Expressway Strategy Alignment}
\label{sec: RL}
\begin{figure}[htbp]
    \centering
    % width 可以设置为 \linewidth (单栏宽度) 或 \textwidth (页面宽度)
    % 如果觉得图太大，可以乘以系数，例如 0.9\linewidth
    \includegraphics[width=1.0\linewidth]{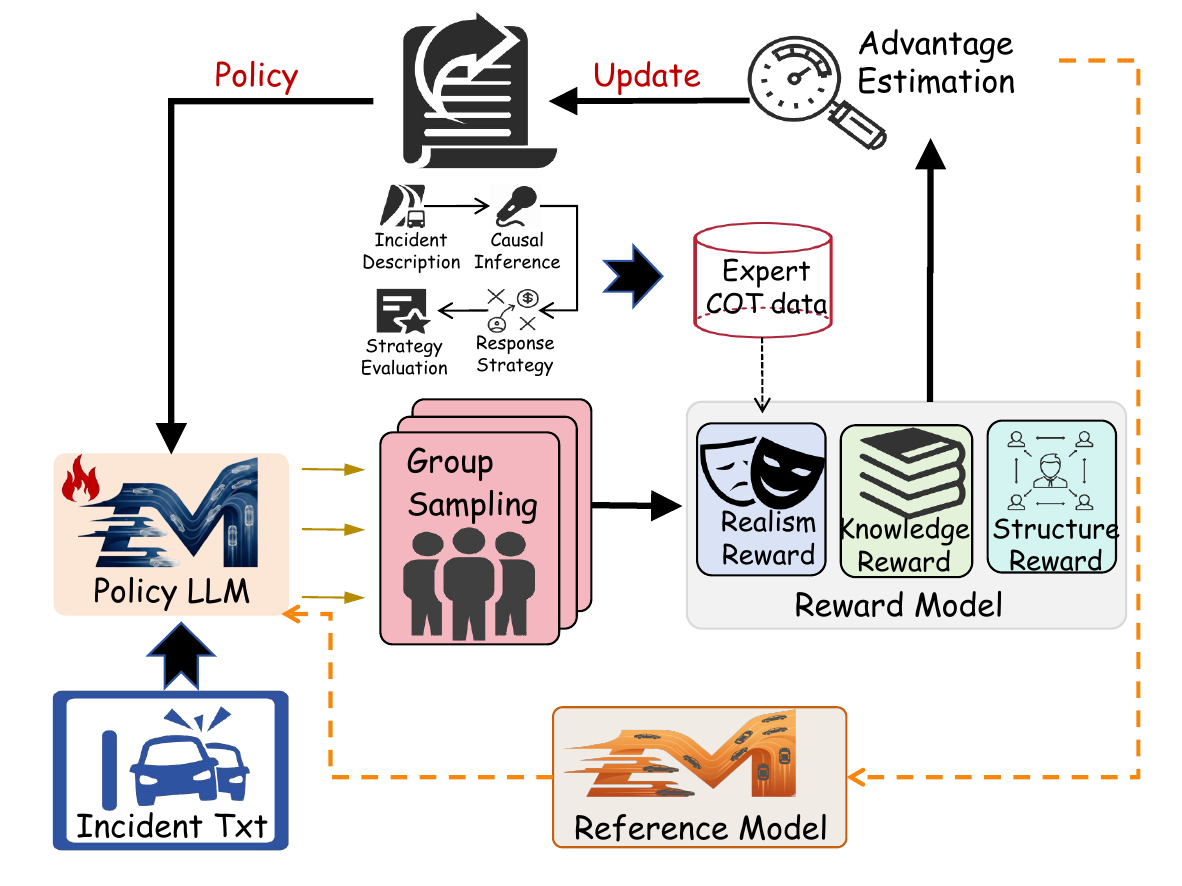}
    \caption{Schematic of the RL-based Reasoning Enhancement.}
    \label{fig:rl_reasoning}
\end{figure}

Although LLMs have acquired fundamental domain cognition through full-parameter pre-training, when dealing with unseen complex expressway accident scenarios, their generated response strategies often fail to establish a complete logical chain from scene analysis to strategy formulation and evaluation. It notably lacks deep logical deduction and fails to ensure strategic optimality. As shown in Figure
\ref{fig:rl_reasoning}, to enforce the "Perception-Analysis-Decision-Reflection" cognitive loop and address its reasoning bottleneck, we leverage a CoT dataset derived from real-world expressway emergency responses and employ the Group Relative Policy Optimization (GRPO) algorithm to mine underlying logical patterns, thereby significantly enhancing the model's reasoning capabilities.

The core mechanism of GRPO involves sampling a group of candidate outputs ${o_1, o_2, \dots, o_G}$ for a given query $q$ and computing gradients by evaluating the relative scores within the group. Here, $q$ is from the set of all possible queries $Q$. The objective function is formulated as follows:

\begin{equation}
    J_{GRPO}(\theta) = \mathbb{E}_{q \sim P(Q), \{o_i\}_{i=1}^G \sim \pi_{\theta_{old}}} \left[ \mathcal{L} \right]
\end{equation}

\begin{equation}
    \mathcal{L} = \frac{1}{G} \sum_{i=1}^G \left( \min \left(r_i A_i, \text{clip}(r_i, 1-\epsilon, 1+\epsilon)A_i \right) - \beta D_{KL}(\pi_\theta || \pi_{ref}) \right)
\end{equation}
where $A_i$ denotes the advantage function. The term $\beta D_{KL}$ acts as a regularization constraint to mitigate catastrophic forgetting during the reinforcement learning process, explicitly ensuring that the model's linguistic generation remains aligned with the standardized traffic terminology acquired during the SFT phase.

%%To compute the advantage $A_i$, GRPO employs a group-wise normalization technique:

%%\begin{equation}
%%A_i = \frac{R_{total}(o_i) - \text{mean}(\mathbf{R}_{group})}{\text{std}(\mathbf{R}_{group}) + \epsilon}
%%\end{equation}

Through this mechanism, the algorithm effectively reduces the variance of gradient estimation, ensuring that the model prioritizes learning the relative superiority of strategies over absolute scores. Therefore, this enables stable policy iteration within the complex reasoning space of traffic incident disposal.

To steer the model towards structured accident response Logic, we design a multi-dimensional reward $R_{total} = \lambda_1 R_{struct} + \lambda_2 R_{know} + \lambda_3 R_{sem}$, comprising three decoupled terms:

\begin{itemize}
    \item \textbf{Structural Integrity ($R_{struct}$):} To enforce the "Perception-Analysis-Decision-Reflection" cognitive loop, we employ a gated counting mechanism. The reward accumulates only if the four stage-specific tags $S_{1..4}$ appear in a strict monotonic order:
    \begin{equation}
        R_{struct} = \left( \sum_{k=1}^4 \mathbb{I}(S_k \in O) \right) \cdot \mathbb{I}\left( \text{idx}(S_1) < \text{idx}(S_2) < \text{idx}(S_3) < \text{idx}(S_4) \right)
    \end{equation}

    \item \textbf{Domain Alignment ($R_{know}$):} We maximize the coverage of stage-specific expert terminology $\mathcal{V}_k$ while penalizing linguistic degradation via a perplexity (PPL) constraint:
    \begin{equation}
        R_{know} = \frac{1}{K} \sum_{k=1}^{K} \omega_k \frac{|S_k \cap \mathcal{V}_k|}{|S_k|} - \eta \cdot \text{ReLU}\big(\text{PPL}(O) - \tau_{ppl}\big)
    \end{equation}

    \item \textbf{Semantic Consistency ($R_{sem}$):} To ensure strategic optimality, we compute the cosine similarity between the model's decision logic and a reference set $\mathcal{D}_{ref}$ containing expert records and teacher traces in the embedding space:
    \begin{equation}
        R_{sem} = \max_{d \in \mathcal{D}_{ref}} \cos \big( \phi(S_2 \oplus S_3), \phi(d) \big)
    \end{equation}
\end{itemize}

The overall algorithm process for RL reasoning alignment is presented in Table \ref{alg:grpo}.  At its core, the study generates decision-making strategies equipped with complete and verifiable expressway emergency response processes. This explicit reasoning trace enhances the interpretability and reliability of the model's outputs. Implementation details regarding the expert vocabulary $\mathcal{V}_k$ and hyperparameters are provided in Appendix \ref{sec:appendix_reward}.

\begin{algorithm}[ht]
\caption{Traffic Incident Strategy Alignment via GRPO}
\label{alg:grpo}
\begin{algorithmic}[1]
\REQUIRE unstructured traffic incident text description $q \sim P(Q)$
\ENSURE optimized traffic disposal policy $\pi^*$
\STATE Initialize policy model $\pi_{\theta}$, reference model $\pi_{ref}$, expert database $\mathcal{D}_{ref}$

\FOR{training epoch $t = 1$ \TO $T$}
    \STATE // Step 1: Candidate Response Sampling
    \STATE Sample $G$ candidate responses from current policy:
    \STATE $\{o_i\}_{i=1}^G \sim \pi_{\theta_{t-1}}(\cdot \mid q)$
    
    \STATE // Step 2: Multi-dimensional Reward Evaluation
    \FOR{each candidate response $o_i$}
        \STATE Compute reward: $r_i = \sum_{j} \lambda_j R_j(o_i)$
        \STATE where $R \in \{R_{struct}, R_{know}, R_{sem}\}$
    \ENDFOR
    
    \STATE // Step 3: Advantage Normalization within Group
    \STATE Compute group mean: $\mu_{\text{group}} = \frac{1}{G} \sum_{i=1}^G r_i$
    \STATE Compute group standard deviation: 
    \STATE $\sigma_{\text{group}} = \sqrt{\frac{1}{G} \sum_{i=1}^G (r_i - \mu_{\text{group}})^2}$
    \FOR{each candidate response $o_i$}
        \STATE Compute advantage: $A_i = \frac{r_i - \mu_{\text{group}}}{\sigma_{\text{group}} + \epsilon}$
    \ENDFOR
    
    \STATE // Step 4: Policy Update via GRPO Objective
    \STATE Update parameters by maximizing GRPO objective:
    \STATE $\theta_t \leftarrow \operatorname{argmax}_{\theta} \; J_{\text{GRPO}}(\theta; A_i, \pi_{\theta_{t-1}}, \pi_{ref})$
\ENDFOR

\STATE \textbf{return} optimized policy $\pi^* \leftarrow \pi_{\theta_T}$
\end{algorithmic}
\end{algorithm}

\subsection{Knowledge Graph-Augmented Retrieval}
\label{sec: RAG}

The static parameters of LLMs cannot capture dynamic information and professional vocabulary. Therefore, this paper constructs a expressway knowledge base to assist LLMs in learning these knowledge. Traffic knowledge is unstructured data so that a graph-based RAG must be adopted to retrieve the knowledge base. This study employs LightRAG \cite{guo2024lightrag} to enhance the ability to update incremental knowledge which introduces a dual-layer retrieval mechanism by constructing a structured graph index

During the indexing phase, an unstructured traffic corpus \( D \) is transformed into a structured and incrementally updatable knowledge graph 
\(\widehat{G} = (\widehat{V}, \widehat{E})\). 
Here, each node \( \widehat{V} \) represents a standardized traffic term, and each edge \( \widehat{E} \) encodes semantic relations. 

This paper designs an entity and relation extraction module to identify the entities specific to the transportation field and their interrelationships. The proposed LLM profiling is the generation of a structured key-value pair \( (K, L) \) for every node \( v \in V \) and edge \( e \in E \), where the key \( K \) serves as a normalized identifier for efficient retrieval, and the value \( L \) is an LLM-generated summary that integrates multi-source definitions and usage contexts. Deduplication merges redundant nodes and edges of different text fragments through semantic similarity comparison. Given a new document \( D' \), its corresponding subgraph \(\widehat{G}' = (\widehat{V}', \widehat{E}')\) is generated independently and merged into the existing graph via set union operations \(\widehat{V} \cup \widehat{V}', \quad \widehat{E} \cup \widehat{E}'\).

In the retrieval and generation phase, a dual-level retrieval paradigm is employed to jointly capture concrete facts and abstract concepts. Given an initial non-standardized response \( \hat{q}_{\text{raw}} \), an LLM first extracts two types of keywords: local terms \( k^{(l)} \) (e.g., "long queue'', "red-green light") and global semantic cues \( k^{(g)} \) (e.g., "traffic congestion", "signal control"). Two parallel retrieval paths are then activated:

Low-level retrieval focuses on exact or near-exact matching by computing the similarity between the embedding of \( k^{(l)} \) and the key embedding of entity nodes:

\begin{equation}
   s_{\text{low}}(v) = \text{sim}\bigl(e(k^{(l)}), e(K_v)\bigr)
\end{equation}
where \( e(\cdot) \) denotes an embedding function, \( \text{sim}(\cdot) \) is typically cosine similarity, and \( K_v, K_e \) denote the retrieval keys of node \( v \) and edge \( e \), respectively.

High-level retrieval operates at the conceptual level by matching \( k^{(g)} \) against topic keys associated with relation edges:

\begin{equation}
   s_{\text{high}}(e) = \text{sim}\bigl(e(k^{(g)}), e(K_e)\bigr)
\end{equation}

All retrieved structured descriptions and associated text snippets are concatenated into a unified context \( C \), which is fed into the LLM to perform term-level replacement along with \( \hat{q}_{\text{raw}} \). The final normalized output \( \hat{q}_{\text{norm}} \) is generated according to the RAG formulation: \(p(\hat{q}_{\text{norm}}\mid \hat{q}_{\text{raw}}, C) \propto \exp\bigl(f_{\text{LLM}}(\hat{q}_{\text{raw}}; C]))\), 
where \( f_{\text{LLM}} \) denotes the LLM internal scoring function.

\subsection{Multimodal Understanding with VPA}
\label{sec: multimodal}

\begin{figure}[htbp]
    \centering
    % width 可以设置为 \linewidth (单栏宽度) 或 \textwidth (页面宽度)
    % 如果觉得图太大，可以乘以系数，例如 0.9\linewidth
    \includegraphics[width=1.0\linewidth]{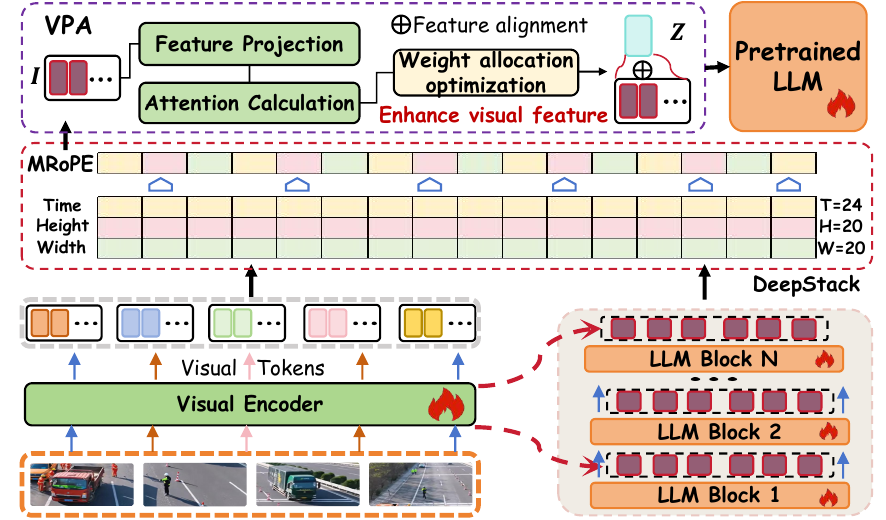}
    \caption{Multimodal Encoding Framework.}
    \label{fig:Cross modal}
\end{figure}

End-to-end Multimodal understanding of expressway is a critical task for expressway supervision. This paper combines a visual encoder with ExpressMind to form a MLLM with video understanding capabilities. In order to enhance the ability of visual feature extraction, this paper introduces a novel visual encoding architecture integrated with a Visual-Prior Alignment (VPA) mechanism, as illustrated in Figure~\ref{fig:Cross modal}. The feature of the visual encoder is \(\mathbf{I} = \{\mathbf{I}_1, \mathbf{I}_2, \dots, \mathbf{I}_t\}, \quad \mathbf{I} \in \mathbb{R}^{N_v \times d_v}\), where \( N_v \) is the total number of visual tokens and \( d_v \) is the dimension of each visual feature vector.

We design a cross-modal projection network and use layer normalization to stabilize the training process as shown in the following equation:

\begin{equation}
\mathbf{H}_v = \text{LayerNorm}\left( \mathbf{W}_2 \cdot \text{GELU}\left( \mathbf{W}_1 \cdot \mathbf{I} + \mathbf{b}_1 \right) + \mathbf{b}_2 \right)
\label{eq:projection}
\end{equation}
where, \({H}_v \in \mathbb{R}^{N_v \times d_{\text{LLM}}}\), \( \mathbf{W}_1 \in \mathbb{R}^{d_v \times d_h} \) and \( \mathbf{W}_2 \in \mathbb{R}^{d_h \times d_{\text{LLM}}} \) are learnable weight matrices, \( \mathbf{b}_1 \) and \( \mathbf{b}_2 \) are bias terms, and \( d_h \) denotes the intermediate hidden dimension. The output dimension \( d_{\text{LLM}} \) matches the hidden size of the LLM.

This paper employs MRoPE to address the issue of uneven frequency allocation in video understanding. It allocates feature channels to the temporal, height, and width axes in a fine‑grained polling manner, ensuring that each positional axis is encoded with a full frequency spectrum ranging from high to low frequencies. Furthermore, this study introduces the DeepStack\cite{meng2024deepstack} mechanism to extract feature maps and perform cross‑layer fusion at different depths, which enables the language decoder to access a complete visual hierarchy spanning from pixel‑level to semantic‑level information.

The visual features may suffer from feature attenuation during the sequence alignment after encoding due to the compression of sequence length. This paper innovatively proposes the VPA mechanism to overcome this problem. It introduces learnable cross-modal attention reweighting to achieve dynamic alignment of visual and language features. VPA explicitly enhances the computational weight of visual tokens in multimodal fusion, establishing a visual-priority inductive preference. Given the projected visual features \( \mathbf{H}_v \) and the text embeddings \( \mathbf{U} \in \mathbb{R}^{L_t \times d_{\text{LLM}}} \) from the instruction prompt, \( L_t \) is the text sequence length. The formula for adjusting the learnable cross-modal attention weights in VPA is as follows:

\begin{equation}
\mathbf{\hat{H}}_v = \text{softmax}\left( \frac{ \mathbf{H}_v \mathbf{W}_p \mathbf{U}^{\top} }{ \sqrt{d_{\text{LLM}}} } \right) \cdot \mathbf{H}_v
\label{eq:vpa}
\end{equation}
where \( \mathbf{W}_p \in \mathbb{R}^{d_{\text{LLM}} \times d_{\text{LLM}}} \) is a learnable projection matrix. This method increases the weight of visual features when aligning visual tokens and text, establishing the inductive bias of visual priors during the feature fusion process.

 The enhanced visual features \( \mathbf{\hat{H}}_v \) are concatenated with the text embeddings \( \mathbf{U} \) to form a unified multimodal input sequence:

\begin{equation}
\mathbf{Z} = \left[ \mathbf{\hat{H}}_v; \mathbf{U} \right] \in \mathbb{R}^{(N_v + L_t) \times d_{\text{LLM}}}
\label{eq:fusion}
\end{equation}

The combined representation \( \mathbf{Z} \) is fed into the LLM and generates natural language description of video, which achieves understanding of traffic scenarios.

\begin{table*}[htbp]
\centering
\caption{Pre-training results on QA tasks.}
\label{tab:pretrain-results-expanded}

% === 布局调整 ===
\footnotesize
% 由于增加了列数，减小列间距以防超宽
\setlength{\tabcolsep}{9pt} 
\renewcommand{\arraystretch}{0.99}

\begin{tabular}{l ccc ccc cc cc}
\toprule
% 表头：Models 垂直居中
\multirow{2}{*}[-2pt]{\textbf{Models}} & \multicolumn{3}{c}{\textbf{MCQ [\%]}} & \multicolumn{3}{c}{\textbf{True/False [\%]}} & \multicolumn{2}{c}{\textbf{Fill-in-the-Blank [\%]}} & \multicolumn{2}{c}{\textbf{Short Answer [\%]}} \\
\cmidrule(lr){2-4} \cmidrule(lr){5-7} \cmidrule(lr){8-9} \cmidrule(lr){10-11}
 & \textbf{Acc} & \textbf{Rec} & \textbf{F1} & \textbf{Acc} & \textbf{Rec} & \textbf{F1} & \textbf{Emb} & \textbf{F1 } & \textbf{GPT-Score} & \textbf{F1} \\
\midrule

% ================= Task 1 =================
\multicolumn{11}{l}{\textit{\textbf{Expressway Laws \& Regulations QA}}} \\
\cmidrule(r){1-11} 
Qwen-32B      & \underline{97.9} & \underline{96.5} & 97.2 & 96.1 & 95.8 & 96.4 & 93.7 & 85.3 & 75.6 & 82.1 \\
Baichuan-32B  & 91.3 & 88.5 & 89.9 & 89.5 & 88.2 & 89.4 & 88.8 & 78.2 & 69.3 & 74.5 \\
DeepSeek-R1-Distill-Qwen-32B   & 96.1 & 96.3 & \underline{97.4} & 96.5 & 96.0 & 96.1 & 92.4 & 84.1 & 76.6 & 82.4 \\
DeepSeek-R1-Distill-Llama-70B      & 97.0 & 96.2 & 96.5 & \underline{97.0} & 95.8 & 96.4 & 95.9 & 88.7 & 85.9 & \underline{87.8} \\
Llama-3.3-70B & 97.5 & 96.0 & 96.8 & 96.5 & \underline{96.2} & \underline{96.8} & \underline{96.6} & \underline{89.4} & \underline{86.1} & 87.2 \\
GLM-4-32B     & 91.7 & 89.4 & 90.5 & 90.1 & 89.5 & 90.3 & 90.4 & 80.5 & 71.4 & 76.4 \\
\textbf{ExpressMind-14B} & \underline{\textbf{98.4}} & \underline{\textbf{97.9}} & \underline{\textbf{98.1}} & \underline{\textbf{98.2}} & \underline{\textbf{97.5}} & \underline{\textbf{98.3}} & \underline{\textbf{97.4}} & \underline{\textbf{90.5}} & \underline{\textbf{86.8}} & \underline{\textbf{88.5}} \\
\midrule

% ================= Task 2 =================
\multicolumn{11}{l}{\textit{\textbf{Smart Expressway Knowledge QA}}} \\
\cmidrule(r){1-11}
Qwen-32B      & 96.5 & 95.2 & 95.8 & 95.5 & 94.8 & 95.7 & 83.1 & 78.5 & 76.4 & 79.5 \\
Baichuan-32B  & 89.7 & 87.5 & 88.6 & 88.2 & 87.0 & 88.1 & 87.7 & 77.5 & 68.4 & 73.8 \\
DeepSeek-R1-Distill-Qwen-32B   & 95.4 & \underline{96.1} & 94.8 & \underline{96.7} & 95.5 & \underline{96.2} & 95.6 & \underline{88.9} & 83.1 & 86.0 \\
DeepSeek-R1-Distill-Llama-70B      & 95.9 & \underline{96.1} & 95.5 & 94.8 & \underline{96.2} & 95.7 & 94.1 & 88.4 & \underline{84.2} & 86.1 \\
Llama-3.3-70B & \underline{\textbf{96.8}} & 95.9 & \underline{96.3} & 95.8 & 95.4 & 96.1 & \underline{95.7} & 88.5 & 84.0 & \underline{86.9} \\
GLM-4-32B     & 90.6 & 88.7 & 89.6 & 90.2 & 89.1 & 90.4 & 89.3 & 79.8 & 70.9 & 75.2 \\
\textbf{ExpressMind-14B} & \underline{96.7} & \underline{\textbf{96.2}} & \underline{\textbf{96.4}} & \underline{\textbf{97.5}} & \underline{\textbf{97.0}} & \underline{\textbf{97.6}} & \underline{\textbf{96.4}} & \underline{\textbf{89.2}} & \underline{\textbf{85.4}} & \underline{\textbf{87.8}} \\
\midrule

% ================= Task 3 =================
\multicolumn{11}{l}{\textit{\textbf{Intelligent Transport System Knowledge QA}}} \\
\cmidrule(r){1-11}
Qwen-32B      & 94.3 & 93.5 & 93.9 & 94.8 & 94.2 & \underline{95.2} & 91.7 & 84.2 & 71.2 & 80.5 \\
Baichuan-32B  & 88.6 & 86.8 & 87.7 & 86.5 & 85.9 & 86.4 & 87.5 & 76.8 & 68.9 & 72.5 \\
DeepSeek-R1-Distill-Qwen-32B   & 93.4 & 94.1 & 93.8 & 92.7 & 92.5 & 93.2 & \underline{95.6} & 87.1 & 83.5 & 85.5 \\
DeepSeek-R1-Distill-Llama-70B      & 93.9 & \underline{94.5} & \underline{94.5} & 92.8 & 93.2 & 92.7 & 94.1 & 87.4 & 84.2 & \underline{86.1} \\
Llama-3.3-70B & \underline{94.9} & 94.1 & \underline{94.5} & \underline{95.0} & \underline{94.5} & 95.1 & 94.2 & \underline{87.6} & \underline{84.7} & 85.8 \\
GLM-4-32B     & 90.1 & 88.5 & 89.3 & 89.4 & 88.8 & 89.7 & 88.6 & 78.9 & 69.7 & 74.1 \\
\textbf{ExpressMind-14B} & \underline{\textbf{95.6}} & \underline{\textbf{95.1}} & \underline{\textbf{95.3}} & \underline{\textbf{96.5}} & \underline{\textbf{96.0}} & \underline{\textbf{96.8}} & \underline{\textbf{95.9}} & \underline{\textbf{88.7}} & \underline{\textbf{84.9}} & \underline{\textbf{86.5}} \\
\bottomrule
\end{tabular}
\end{table*}

\section{Experiments}
\subsection{Dataset}
This study released a comprehensive open-source expressway dataset comprising four specialized sub-datasets and a standardized benchmark, for the training and evaluation of \textbf{ExpressMind}, focusing on multi-modal capabilities such as expressway domain knowledge understanding, incident response strategy reasoning, video understanding and incident detection.

\textbf{1: Express-Insight} contains over 7 million tokens of high-quality text where the content serves as a domain-specific corpus for unsupervised pre-training. The dataset is derived from web-crawled resources including traffic laws, expressway policy documents, and theoretical books on Smart expressways and Intelligent Transportation Systems.

\textbf{2: Express-QA} contains over 870,000 samples where QA pairs are obtained through a quality-aware generation and refinement process using DeepSeek-V3 with designed prompts.

\textbf{3: Express-IncidentCoT} contains 1,786 incident response strategy Chain-of-Thought samples derived from real-world incident reports of Shandong Expressway, structured into a four-stage cognitive chain: [Incident Description] $\to$ [Causal Inference] $\to$ [Response Strategy Formulation] $\to$ [Strategy Evaluation].

\textbf{4: Express-VQA} a multi-modal dataset for expressway visual reasoning, integrating 1627 surveillance videos from expressways in Shandong and Guangdong, China. The average duration of the videos exceeds 2 minutes. There are over 3,200 pairs of VQA pairs. There are also 12 sets of surveillance video from Tianjin that cover two consecutive days. The data is collected across 70 roads at 1920×1080 resolution, encompassing diverse times and weather conditions to evaluate model robustness against seven core traffic anomalies such as accidents, congestion, and construction.

\subsection{Experiment Setup}

The experimental environment is configured with the following specifications: All experimental workflows---including model training, testing, and inference---were executed on a server node equipped with 8 NVIDIA H20 GPUs. The software environment is built upon Python 3.10+, PyTorch 2.4.0+, and CUDA 12.4+. The transformers, tokenizers, torchvision, and opencv-python libraries are employed for processing text, model, and image data, respectively. Stable versions of DeepSpeed, Accelerate, PEFT, and Flash-Attention 2 are utilized to facilitate efficient distributed fine-tuning, thereby ensuring high training efficiency and stability. Under this configuration, the end-to-end training of the framework required approximately 700 gpu hours. In particular, the hyperparameter configurations for each training stage are detailed in Appendix \ref{sec:Metircs}. 

\subsection{Quantitative Results}
\subsubsection{Pre-training Results.}

\begin{figure}[htbp]
    \centering
    % width 可以设置为 \linewidth (单栏宽度) 或 \textwidth (页面宽度)
    % 如果觉得图太大，可以乘以系数，例如 0.9\linewidth
    \includegraphics[width=\linewidth]{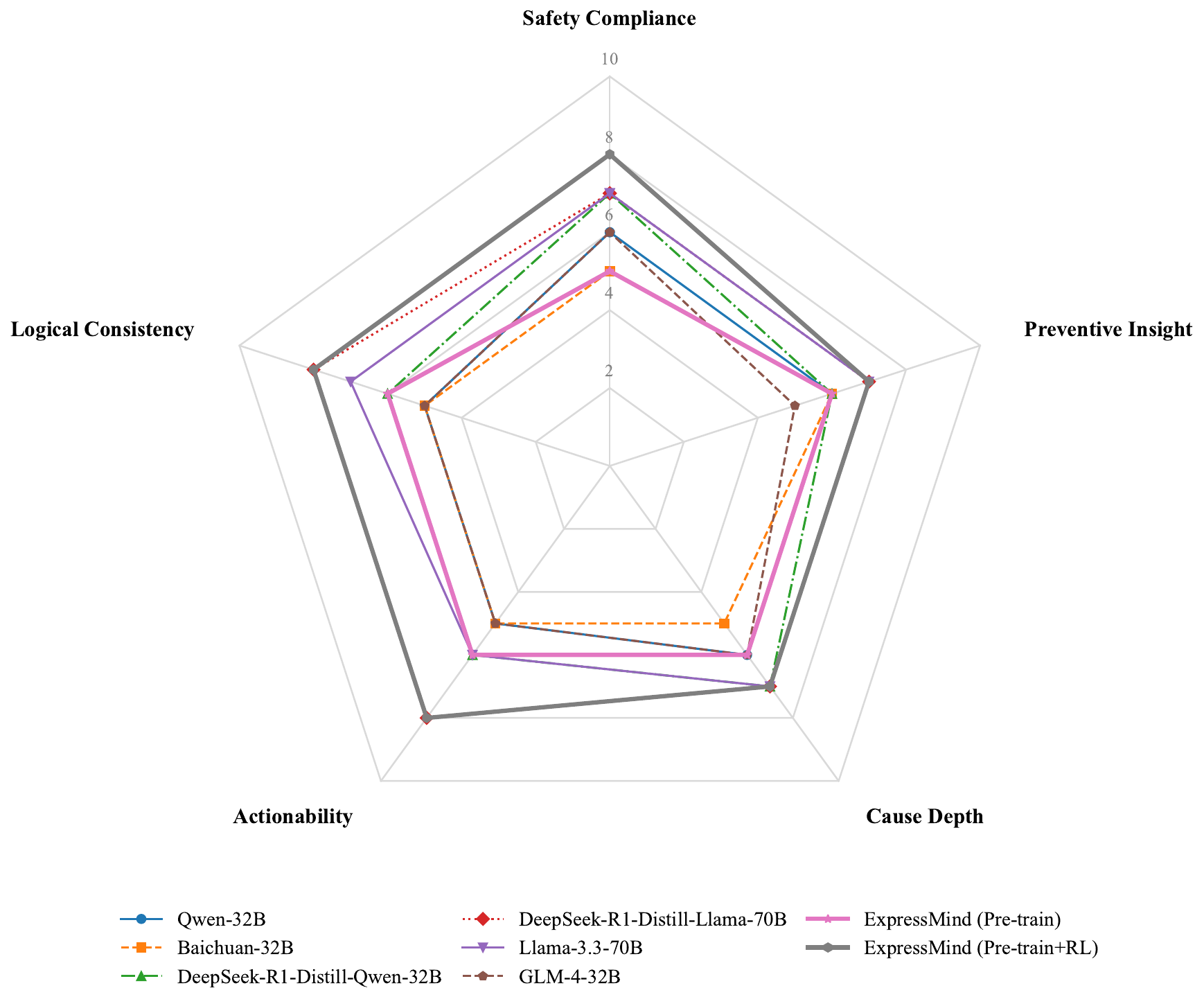}
    \caption{Performance of ExpressMind with RL Alignment.}
    \label{fig:rl_reasoning_results}
\end{figure}

The results presented in Table \ref{tab:pretrain-results-expanded} summarize the performance of ExpressMind across three specialized tasks (totaling 20,000 test questions): Expressway Laws \& Regulations QA, Smart Expressway Knowledge QA, and Intelligent Transportation System Knowledge QA. For each task, We used a set of multi‑dimensional metrics, including Accuracy, F1‑Score, Embedding Similarity, and GPT‑Score, to benchmark our model againstestablished open-source baseline models such as Qwen-32B \cite{yang2025qwen3}, Llama-3.3-70B \cite{grattafiori2024llama}, and the DeepSeek-R1-Distill series \cite{guo2025deepseek}. Despite its specialized focus, ExpressMind consistently outperforms these LLM across all evaluation dimensions. Notably, in the \textit{Expressway Laws \& Regulations QA} task, our model achieves a peak MCQ accuracy of $98.4\%$ and a Short Answer F1-score of $88.5\%$, surpassing the strongest baseline, Llama-3.3-70B, by a significant margin. Furthermore, regarding the GPT-Score, which evaluates deep semantic understanding, ExpressMind maintains a high average of $85.7\%$, demonstrating superior competitiveness against reasoning-distilled models like DeepSeek-R1-Distill-Llama-70B. These results emphasize ExpressMind's expert-level proficiency and its ability to provide precise, logically coherent responses within the specialized expressway transportation domain.

\begin{figure}[htbp]
    \centering
    % width 可以设置为 \linewidth (单栏宽度) 或 \textwidth (页面宽度)
    % 如果觉得图太大，可以乘以系数，例如 0.9\linewidth
    \includegraphics[width=1.0\linewidth]{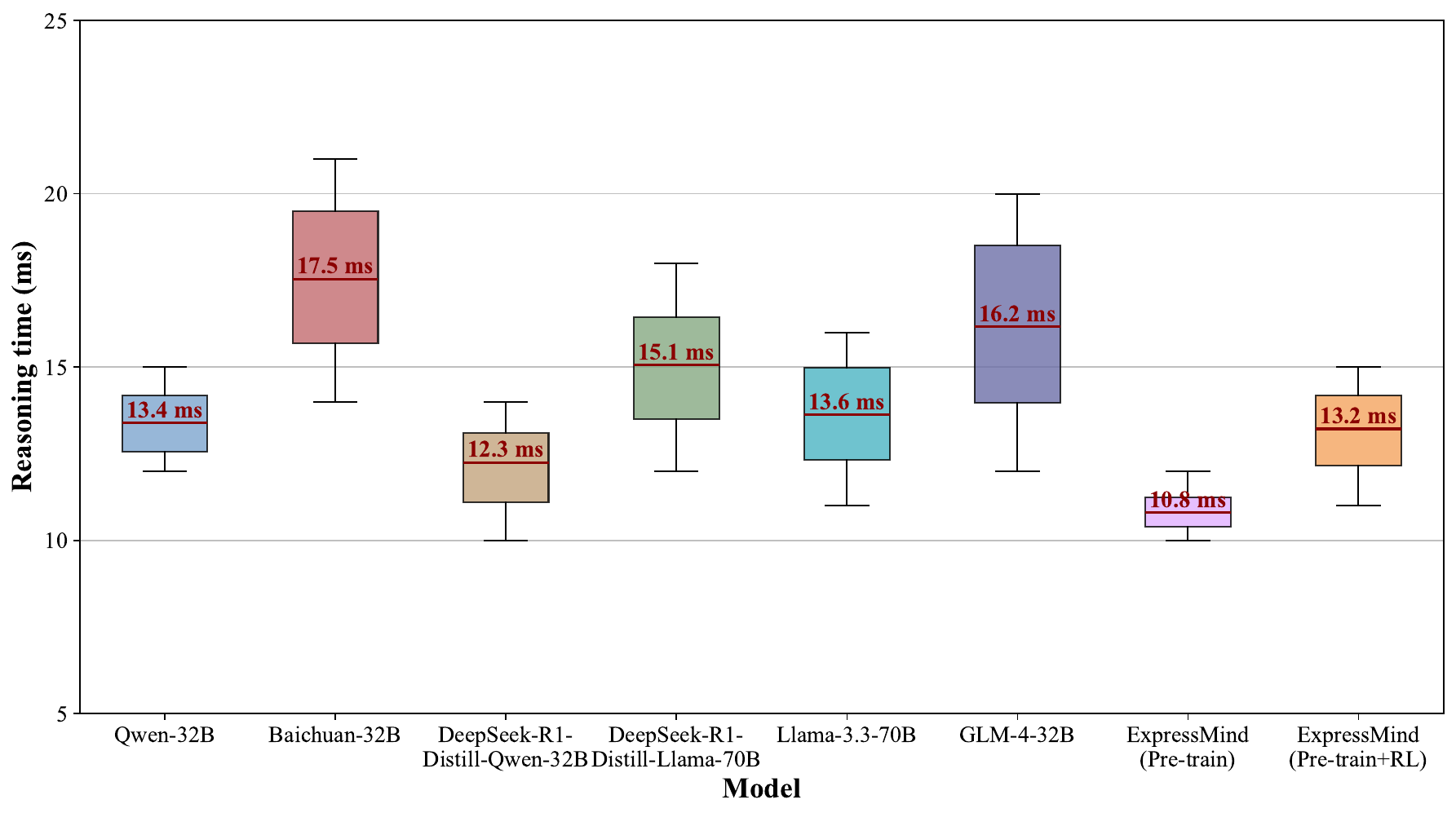}
    \caption{The Reasoning Time of the RL Alignment.}
    \label{fig:rl_reasoning_time}
\end{figure}

\begin{figure*}[t] % 加 * 号表示跨双栏，[t] 表示优先置顶
  \centering % 图片居中
  \includegraphics[width=\textwidth]{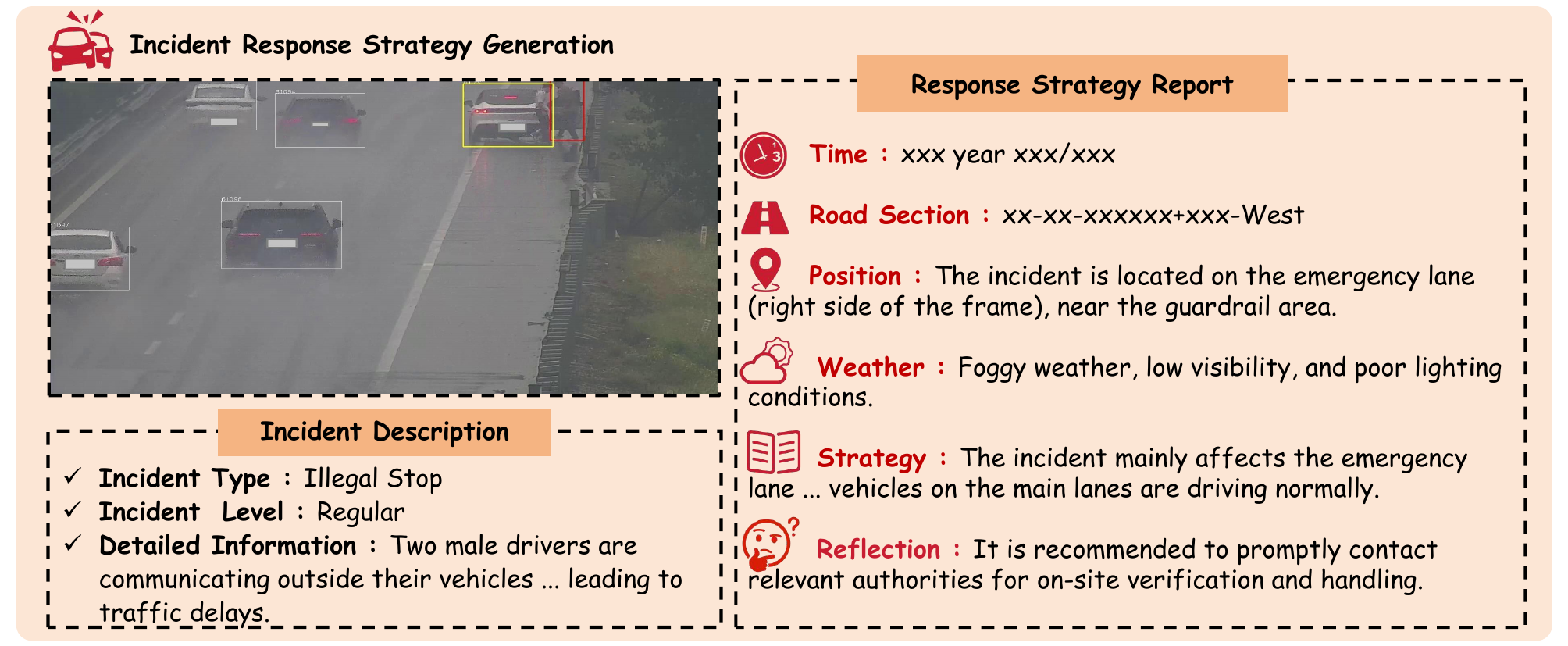}
  \caption{An Example of Incident Response Strategy Generation.}
  \label{fig:Incident_Response}
\end{figure*}

\subsubsection{RL Alignment Results. }

The results presented in Figure \ref{fig:rl_reasoning_results} indicate that ExpressMind (Pretrain+RL) consistently outperforms existing generalist baseline methods in the specialized task of expressway incident management strategy generation on five metrics, detailed in appendix \ref{sec:Metircs}. Specifically, in domain-critical metrics such as Safety Compliance and Actionability, ExpressMind (Pretrain+RL) achieves scores in the range of 8.0–9.0, which is notably higher than general baselines like Llama-3.3-70B and Qwen-32B. The ablation study between the two ExpressMind configurations underscores the decisive impact of RL alignment: the base pretrained model without RL tuning yields the weakest performance, whereas its RL-aligned counterpart demonstrates a significant improvement. This enhancement is directly attributable to the model's better-aligned chain-of-thought reasoning with the required procedural knowledge.
As shown in Figure \ref{fig:rl_reasoning_time}, ExpressMind (Pre-train+RL) demonstrates exceptional deterministic performance in terms of reasoning efficiency. Experimental results indicate that its average inference latency is reduced to 13.2 ms, achieving a 24.6\% acceleration compared to models like Baichuan-32B. More importantly, the tightly clustered distribution in the box plot reveals minimal latency jitter during instruction processing. This combination of low latency and high stability ensures reliable performance for time-sensitive applications, such as millisecond-level response requirements in smart expressway scenarios.

\begin{figure}[thbp]
    \centering
    % width 可以设置为 \linewidth (单栏宽度) 或 \textwidth (页面宽度)
    % 如果觉得图太大，可以乘以系数，例如 0.9\linewidth
    \includegraphics[width=1.0\linewidth]{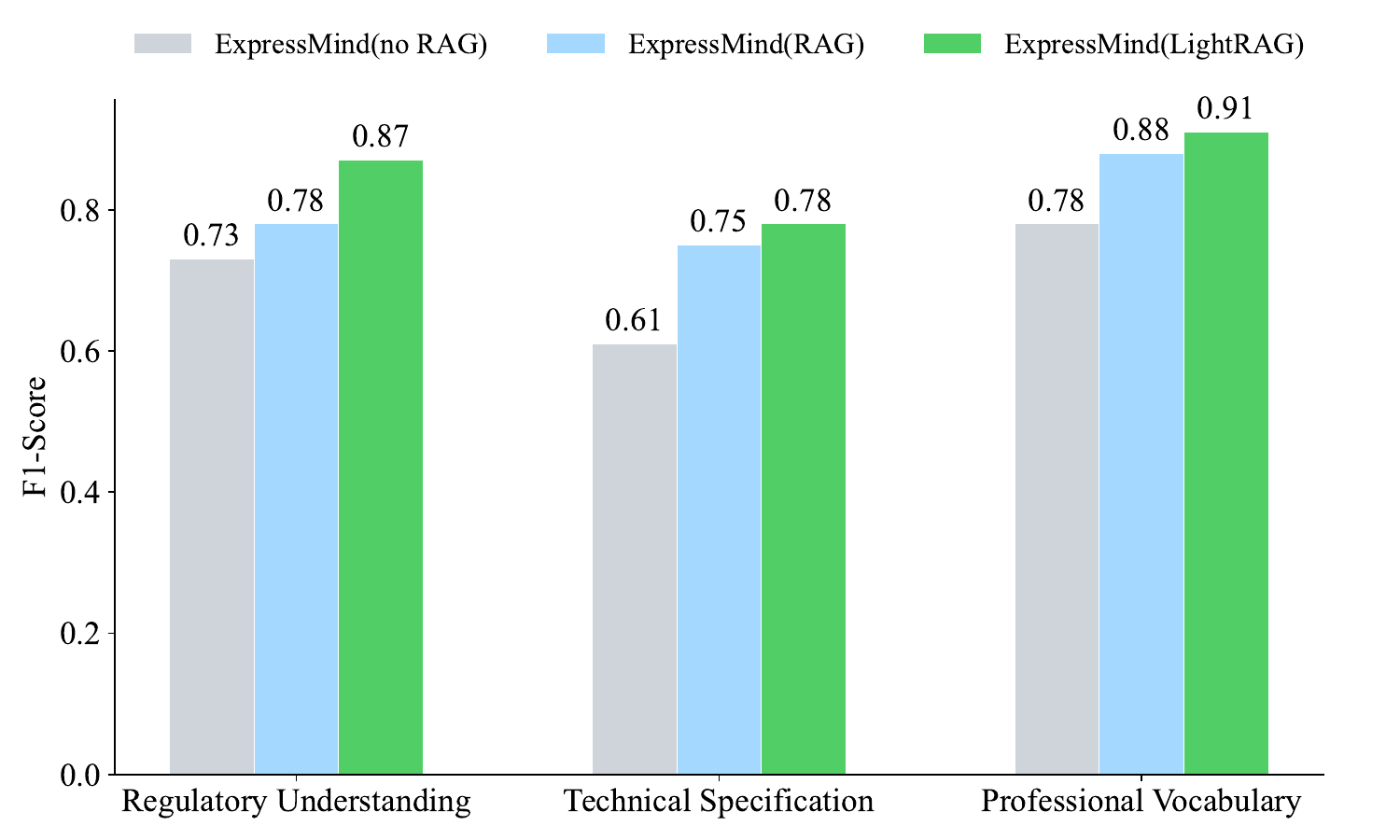}
    \caption{The Ablation Experiment of RAG.}
    \label{fig:RAG}
\end{figure}

\subsubsection{Knowledge Retrieval Ablation Study. }

This section verifies the improvement in the generation capability of ExpressMind resulting from the use of the expressway knowledge base. We leverage LightRAG to query this repository. The evaluation dataset comprises 200 queries addressing complex traffic regulation comprehension and 100 queries focused on technical specifications.

We employ the F1-Score to evaluate the factual accuracy of the generated responses. As illustrated in Figure \ref{fig:RAG}, the experimental results demonstrate that ExpressMind exhibits exceptional reasoning capabilities and robust retrieval performance. The Expressway knowledge base can increase the occurrence probability of professional vocabularies by 16.7\%.

\subsubsection{Scene Understanding Comparison. }
Built upon the ExpressMind backbone and a sophisticated cross-modal encoder, the MLLM, ExpressMind-VL, demonstrates exceptional proficiency in understanding traffic videos. In this study, we evaluate its performance against a range of leading MLLMs, including VideoLLaMA 3 \cite{zhang2025videollama}, MiniCPM-V 4.5 \cite{yu2025minicpm}, InternVL 3.5 \cite{wang2025internvl3}, and Qwen3-VL \cite{bai2025qwen3vltechnicalreport}.

\begin{figure}[t] % 如果是双栏论文想占全宽，请用 figure*；单栏或仅占一栏用 figure
    \centering

    % 左侧子图
    \begin{subfigure}{0.48\linewidth} 
        \centering
        \includegraphics[width=\linewidth]{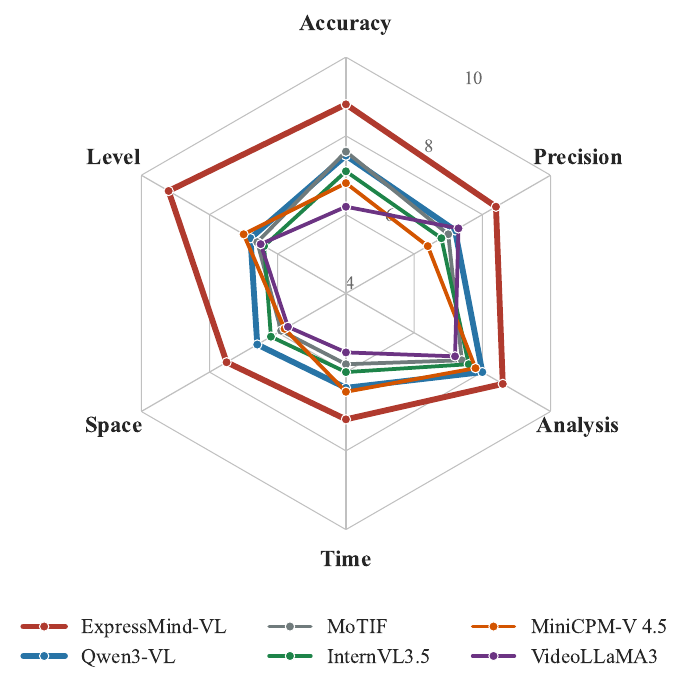}
        \caption{\raggedright Shandong Expressway}
        \label{fig:radar_solid}
    \end{subfigure}
    \hfill % 在两个子图之间插入弹性间距，将它们推向两端
    % 右侧子图
    \begin{subfigure}{0.48\linewidth}
        \centering
        \includegraphics[width=\linewidth]{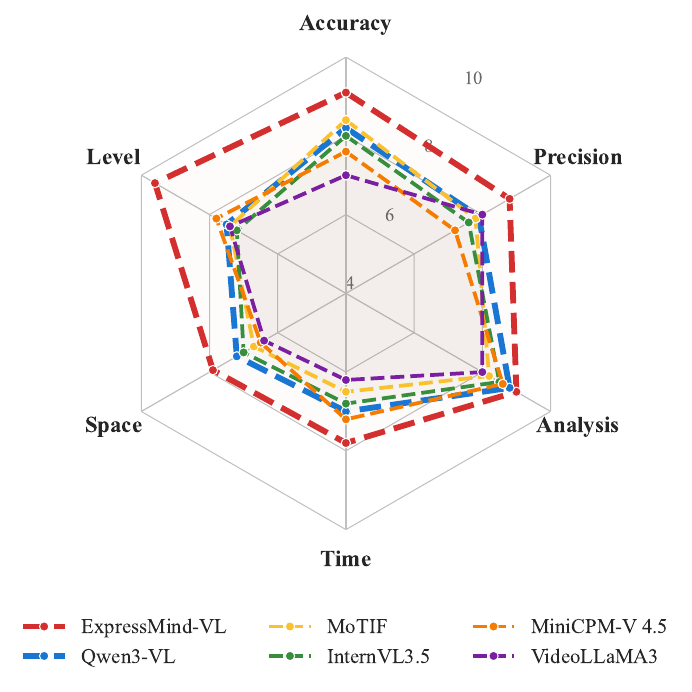}
        \caption{\raggedright Guangdong Expressway}
        \label{fig:radar_no_shadow}
    \end{subfigure}

    \caption{Results of Traffic Incident Detection.}
    \label{fig:radar_comparison_total}
\end{figure}

To quantitatively evaluate the quality of the generated descriptions, we employ four standard automated metrics: BLEU-4 \cite{post2018clarityreportingbleuscores}, ROUGE-L \cite{lin2004rouge}, CIDEr \cite{vedantam2015cider}, and BERTScore \cite{zhang2019bertscore}. Each metric assesses a distinct dimension of linguistic fidelity, ranging from lexical overlap to semantic similarity. To ensure reproducibility, we conducted a comprehensive evaluation using a dataset of 670 expressway surveillance videos input into all comparative models.

As presented in Table \ref{tab:vl_results}, the experimental results demonstrate that ExpressMind-VL significantly outperforms other generalist models in the descriptive accuracy of expressway scenes. The model exhibits a superior capability to interpret complex traffic scenarios, validating the effectiveness of our domain-specific multimodal alignment.

\begin{figure}[t] % 如果是双栏论文想占全宽，请用 figure*；单栏或仅占一栏用 figure
    \centering

    % 左侧子图
    \begin{subfigure}{0.49\linewidth}
        \centering
        \includegraphics[width=\linewidth]{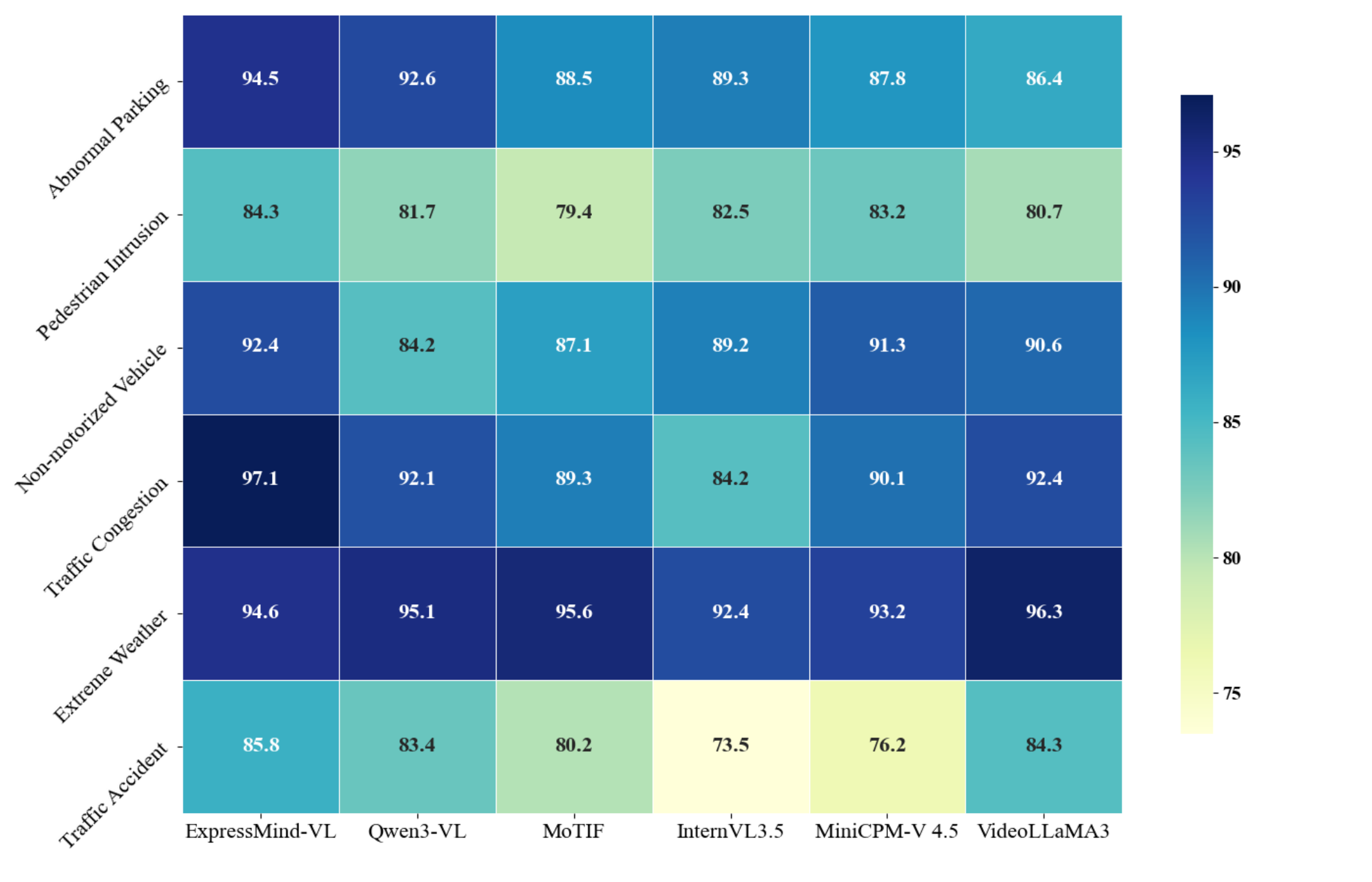}
        \caption{\raggedright Detection Accuracy}
        \label{fig:radar_solid}
    \end{subfigure}
    %\hfill % 在两个子图之间插入弹性间距，将它们推向两端
    % 右侧子图
    \begin{subfigure}{0.49\linewidth}
        \centering
        \includegraphics[width=\linewidth]{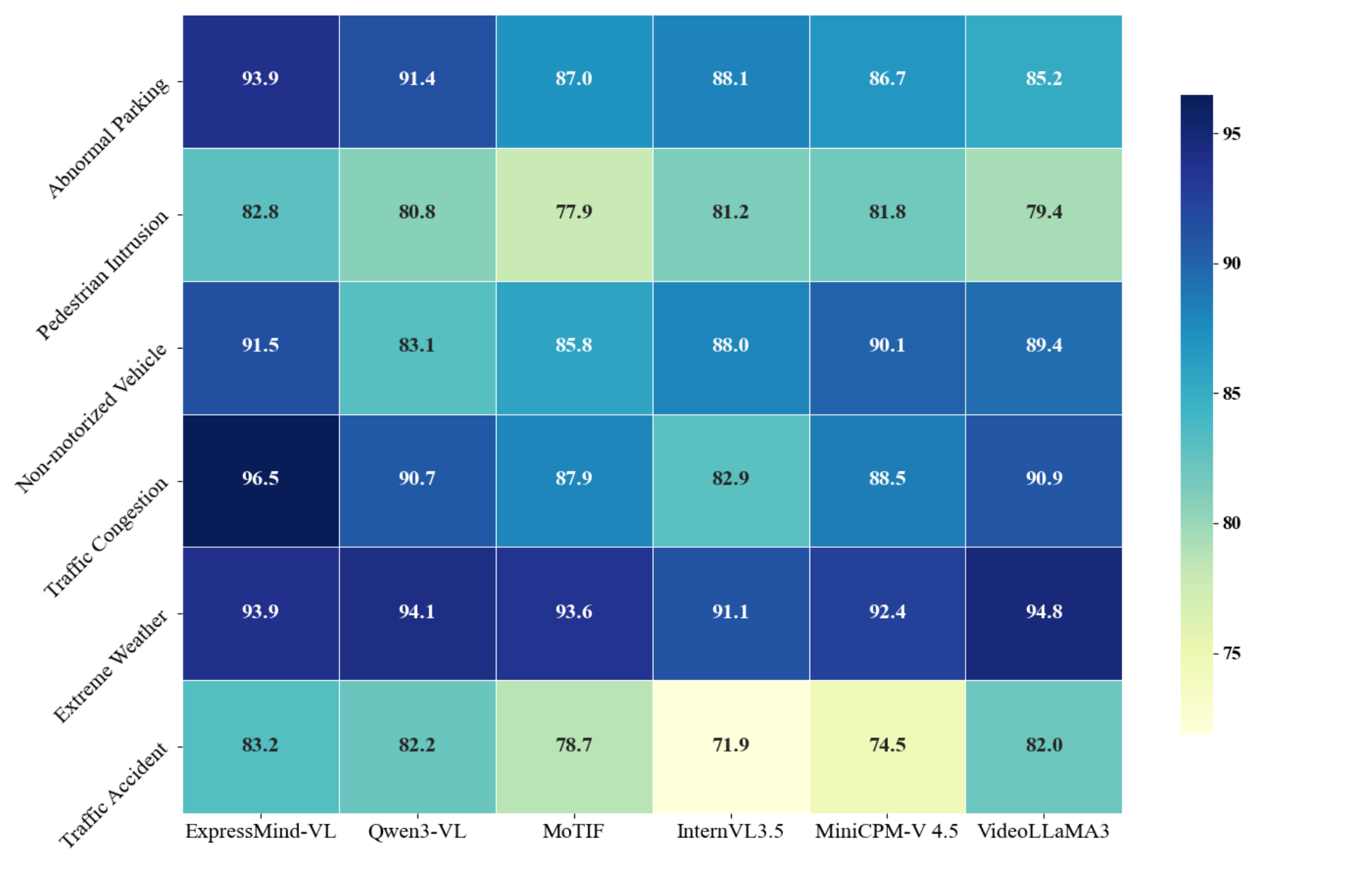}
        \caption{\raggedright Detection Recall}
        \label{fig:radar_no_shadow}
    \end{subfigure}

    \caption{Results of Traffic Incident Detection.}
    \label{fig:heat}
\end{figure}

\begin{table}[!ht]
\centering
\caption{Automated evaluation of quantitative results.}
\label{tab:vl_results}

% \resizebox{宽度}{高度}{内容}
% \linewidth 表示适应当前行宽
% ! 表示高度自动适应，保持字体长宽比不变
\resizebox{\linewidth}{!}{%
    \begin{tabular}{llcccc}
    \toprule
    Model & LLM & BLEU-4 & ROUGE-L & CIDEr & BERTScore \\
    \midrule
    MoTIF & LLaMA2-7B & 82.59 & 87.45 & 69.95 & 88.43 \\
    VideoLLaMA3 & Qwen 2.5-7B & 79.63 & 85.32 & 67.42 & 85.33 \\
    MiniCPM-V 4.5 & LLaMA 3-8B & 81.98 & 87.13 & 69.76 & 88.04 \\
    InternVL3.5 & Qwen 3-38B & 84.53 & 88.94 & 71.67 & 88.97 \\
    Qwen3-VL & Qwen 3-32B & 84.85 & \textbf{89.30} & 72.96 & 89.18 \\
    ExpressMind-VL & ExpressMind-14B & \textbf{85.24} & 89.25 & \textbf{73.36} & \textbf{89.28} \\
    \bottomrule
    \end{tabular}% <--- 这里加个注释符防止产生多余空格
}
\end{table}

Accurate detection and comprehensive analysis of traffic incidents are paramount for intelligent expressway operation. To rigorously evaluate these capabilities, we benchmark ExpressMind-VL against the high-performing Qwen3-VL-32B using a curated dataset of 200 traffic incident videos. The quantitative evaluation employs a multi-dimensional metric suite: Accuracy and Precision for event classification, and F1-scores to assess the semantic fidelity of descriptive texts concerning event severity (Level), causal analysis, spatiotemporal context (Time \& Space), and congestion status (Queue). As illustrated in Figure \ref{fig:radar_comparison_total}, experimental results demonstrate that ExpressMind-VL exhibits significantly superior recognition and reasoning capabilities in traffic incident detection and analysis compared to the baseline. Furthermore, a representative example of Incident Response Strategy Generation, which translates these analytical insights into actionable decisions, is visualized in Figure \ref{fig:heat}.

\begin{figure}[t] % 如果是双栏论文想占全宽，请用 figure*；单栏或仅占一栏用 figure
    \centering

    % 左侧子图
    \begin{subfigure}{0.48\linewidth} 
        \centering
        \includegraphics[width=\linewidth]{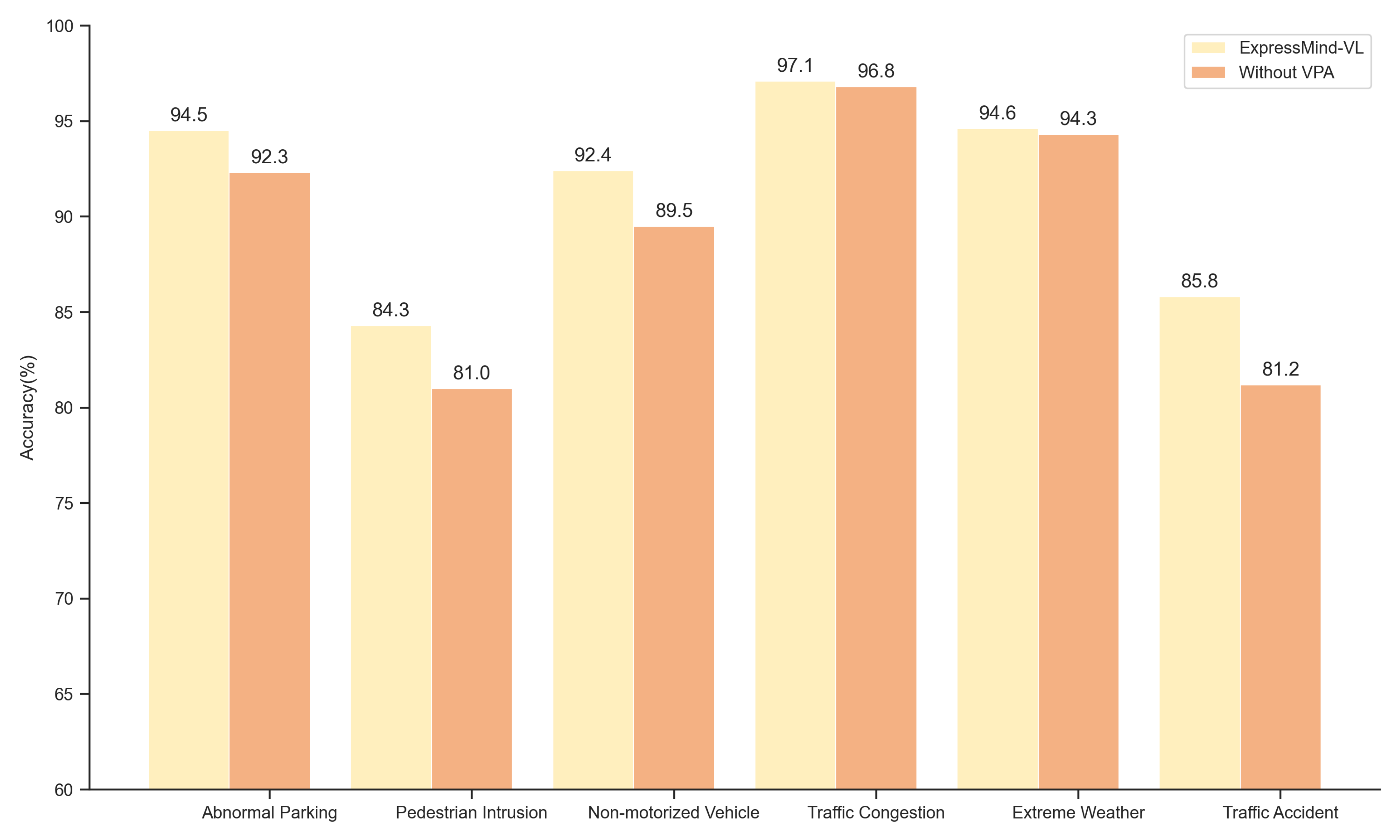}
        \caption{\raggedright Accuracy}
        \label{fig:radar_solid}
    \end{subfigure}
    \hfill % 在两个子图之间插入弹性间距，将它们推向两端
    % 右侧子图
    \begin{subfigure}{0.48\linewidth}
        \centering
        \includegraphics[width=\linewidth]{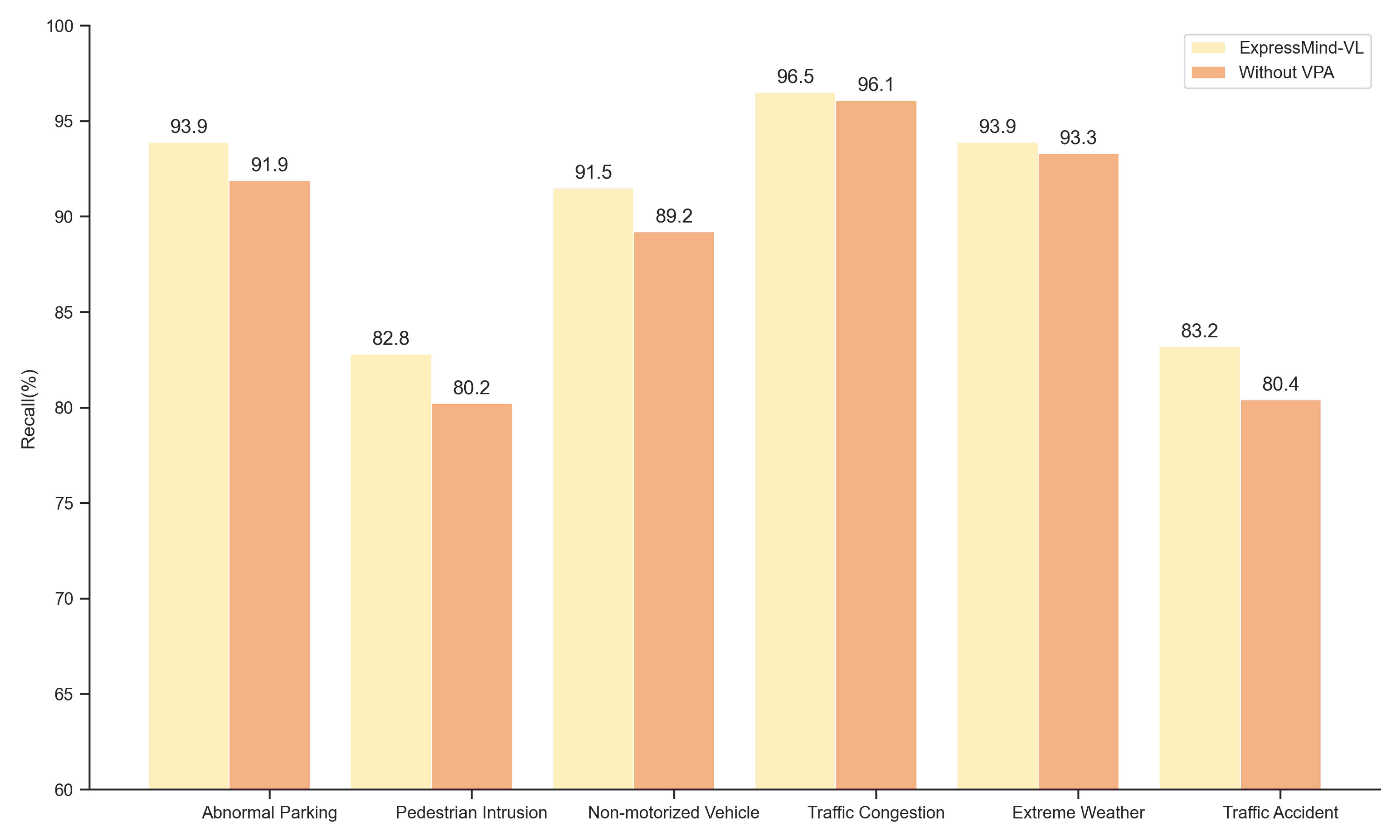}
        \caption{\raggedright Recall}
        \label{fig:radar_no_shadow}
    \end{subfigure}

    \caption{Ablation Results of VPA.}
    \label{fig:Ablation}
\end{figure}

To evaluate the practical performance of ExpressMind-VL in real-world expressway scenarios, we assessed its detection capability for six core types of traffic incidents on the Express-VQA dataset. As shown in Figure \ref{fig:heat}, the accuracy and recall rates of ExpressMind-VL have both exceeded 90\% across all tasks, including abnormal parking, pedestrian intrusion, non-motorized vehicle detection, traffic congestion, extreme weather, and traffic incidents. The high precision and recall of ExpressMind-VL can be attributed to the multi-level technical strategies integrated during its development. Pre-training on high-quality video-text pairs established a robust foundation for cross-modal semantic alignment. As shown in Figure \ref{fig:Ablation}, the VPA mechanism explicitly enhances the contribution of visual features, ensuring the dominance of dynamic visual cues in the reasoning process. Furthermore, standardized prompt engineering reformulates the detection task into a structured text generation problem, enabling the model to naturally incorporate prior knowledge such as traffic rules into logical reasoning.

\subsection{Deployment Analysis}
The proposed ExpressMind has already been deployed in the Shandong Expressway Cloud Brain system. It demonstrates domain-specific comprehension in professional knowledge question-answering and the generation of emergency response strategies for traffic incidents. Furthermore, addressing user needs for customized ExpressMind-VL functionality, we have developed an ExpressMind-VL-based expressway incident monitoring and management platform for both Shandong and Zhejiang expressways. This study designs standardized prompt engineering and a video stream detection mechanism, enabling the model to retain 10 seconds of video upon detecting a traffic incident while simultaneously generating a structured incident analysis report. Through this approach, the system can automatically identify traffic incidents, analyze Traffic scenes, and produce response plans based on real-time conditions. Ultimately, ExpressMind-VL achieves a fully autonomous "perception-analysis-decision" processing pipeline for expressway traffic incidents, serving as an intelligent central hub of expressway. The demonstration of the system application is described in the Appendix \ref{sec: App}.

\section{Conclusion}

This study introduces ExpressMind, the first domain-specific MLLM designed for expressway scenarios. It is built through multiple technical innovations: a two-stage pretraining paradigm for domain knowledge internalization, a GRPO-enhanced RL framework for safety-critical reasoning alignment, a graph-augmented RAG mechanism for real-time spatiotemporal knowledge retrieval, and a VPA multimodal alignment module for deep video understanding. To support this work, we have open-sourced the first training dataset covering domain knowledge, incident CoT strategy reasoning, and multimodal incident detection VQA. The ExpressMind has been applied in top-tier expressway groups, serving as a representative application case of large models in the expressway domain.

In future work, we will focus on three key improvements: enhancing multimodal spatiotemporal reasoning for dynamic scenario understanding, strengthening chain-of-thought alignment in long-text analysis, and advancing model lightweighting for edge deployment. 

%%
%%\section{Acknowledgments}

%%Identification of funding sources and other support, and thanks to
%%individuals and groups that assisted in the research and the
%%preparation of the work should be included in an acknowledgment
%%section, which is placed just before the reference section in your
%%document.

%%

%\begin{acks}
%This work is supported by the Transportation Science and Technology Plan Project of Shandong Transportation Department (Grant No. 2025BAI20).
%\end{acks}

% =========================================================
% =========================================================
\clearpage
\bibliographystyle{ACM-Reference-Format}
\bibliography{main}


\begin{thebibliography}{40}

%%% the Web address.

\ifx \showCODEN    \undefined \def \showCODEN     #1{\unskip}     \fi
\ifx \showISBNx    \undefined \def \showISBNx     #1{\unskip}     \fi
\ifx \showISBNxiii \undefined \def \showISBNxiii  #1{\unskip}     \fi
\ifx \showISSN     \undefined \def \showISSN      #1{\unskip}     \fi
\ifx \showLCCN     \undefined \def \showLCCN      #1{\unskip}     \fi
\ifx \shownote     \undefined \def \shownote      #1{#1}          \fi
\ifx \showarticletitle \undefined \def \showarticletitle #1{#1}   \fi
\ifx \showURL      \undefined \def \showURL       {\relax}        \fi
% invisible to TeX
\providecommand\bibfield[2]{#2}
\providecommand\bibinfo[2]{#2}
\providecommand\natexlab[1]{#1}
\providecommand\showeprint[2][]{arXiv:#2}

\bibitem[Achiam et~al\mbox{.}(2023)]%
        {achiam2023gpt}
\bibfield{author}{\bibinfo{person}{Josh Achiam}, \bibinfo{person}{Steven Adler}, \bibinfo{person}{Sandhini Agarwal}, \bibinfo{person}{Lama Ahmad}, \bibinfo{person}{Ilge Akkaya}, \bibinfo{person}{Florencia~Leoni Aleman}, \bibinfo{person}{Diogo Almeida}, \bibinfo{person}{Janko Altenschmidt}, \bibinfo{person}{Sam Altman}, \bibinfo{person}{Shyamal Anadkat}, {et~al\mbox{.}}} \bibinfo{year}{2023}\natexlab{}.
\newblock \showarticletitle{Gpt-4 technical report}.
\newblock \bibinfo{journal}{\emph{arXiv preprint arXiv:2303.08774}} (\bibinfo{year}{2023}).
\newblock


\bibitem[Arefeen et~al\mbox{.}(2024)]%
        {arefeen2024trafficlens}
\bibfield{author}{\bibinfo{person}{Md~Adnan Arefeen}, \bibinfo{person}{Biplob Debnath}, {and} \bibinfo{person}{Srimat Chakradhar}.} \bibinfo{year}{2024}\natexlab{}.
\newblock \showarticletitle{TrafficLens: Multi-Camera Traffic Video Analysis Using LLMs}. In \bibinfo{booktitle}{\emph{2024 IEEE 27th International Conference on Intelligent Transportation Systems (ITSC)}}. IEEE, \bibinfo{pages}{3974--3981}.
\newblock


\bibitem[Bai et~al\mbox{.}(2025a)]%
        {bai2025qwen3vltechnicalreport}
\bibfield{author}{\bibinfo{person}{Shuai Bai}, \bibinfo{person}{Yuxuan Cai}, \bibinfo{person}{Ruizhe Chen}, \bibinfo{person}{Keqin Chen}, \bibinfo{person}{Xionghui Chen}, \bibinfo{person}{Zesen Cheng}, \bibinfo{person}{Lianghao Deng}, \bibinfo{person}{Wei Ding}, \bibinfo{person}{Chang Gao}, \bibinfo{person}{Chunjiang Ge}, \bibinfo{person}{Wenbin Ge}, \bibinfo{person}{Zhifang Guo}, \bibinfo{person}{Qidong Huang}, \bibinfo{person}{Jie Huang}, \bibinfo{person}{Fei Huang}, \bibinfo{person}{Binyuan Hui}, \bibinfo{person}{Shutong Jiang}, \bibinfo{person}{Zhaohai Li}, \bibinfo{person}{Mingsheng Li}, \bibinfo{person}{Mei Li}, \bibinfo{person}{Kaixin Li}, \bibinfo{person}{Zicheng Lin}, \bibinfo{person}{Junyang Lin}, \bibinfo{person}{Xuejing Liu}, \bibinfo{person}{Jiawei Liu}, \bibinfo{person}{Chenglong Liu}, \bibinfo{person}{Yang Liu}, \bibinfo{person}{Dayiheng Liu}, \bibinfo{person}{Shixuan Liu}, \bibinfo{person}{Dunjie Lu}, \bibinfo{person}{Ruilin Luo}, \bibinfo{person}{Chenxu Lv}, \bibinfo{person}{Rui Men},
  \bibinfo{person}{Lingchen Meng}, \bibinfo{person}{Xuancheng Ren}, \bibinfo{person}{Xingzhang Ren}, \bibinfo{person}{Sibo Song}, \bibinfo{person}{Yuchong Sun}, \bibinfo{person}{Jun Tang}, \bibinfo{person}{Jianhong Tu}, \bibinfo{person}{Jianqiang Wan}, \bibinfo{person}{Peng Wang}, \bibinfo{person}{Pengfei Wang}, \bibinfo{person}{Qiuyue Wang}, \bibinfo{person}{Yuxuan Wang}, \bibinfo{person}{Tianbao Xie}, \bibinfo{person}{Yiheng Xu}, \bibinfo{person}{Haiyang Xu}, \bibinfo{person}{Jin Xu}, \bibinfo{person}{Zhibo Yang}, \bibinfo{person}{Mingkun Yang}, \bibinfo{person}{Jianxin Yang}, \bibinfo{person}{An Yang}, \bibinfo{person}{Bowen Yu}, \bibinfo{person}{Fei Zhang}, \bibinfo{person}{Hang Zhang}, \bibinfo{person}{Xi Zhang}, \bibinfo{person}{Bo Zheng}, \bibinfo{person}{Humen Zhong}, \bibinfo{person}{Jingren Zhou}, \bibinfo{person}{Fan Zhou}, \bibinfo{person}{Jing Zhou}, \bibinfo{person}{Yuanzhi Zhu}, {and} \bibinfo{person}{Ke Zhu}.} \bibinfo{year}{2025}\natexlab{a}.
\newblock \bibinfo{title}{Qwen3-VL Technical Report}.
\newblock
\showeprint[arxiv]{2511.21631}~[cs.CV]
\urldef\tempurl%
\url{https://arxiv.org/abs/2511.21631}
\showURL{%
\tempurl}


\bibitem[Bai et~al\mbox{.}(2025b)]%
        {bai2025qwen2}
\bibfield{author}{\bibinfo{person}{Shuai Bai}, \bibinfo{person}{Keqin Chen}, \bibinfo{person}{Xuejing Liu}, \bibinfo{person}{Jialin Wang}, \bibinfo{person}{Wenbin Ge}, \bibinfo{person}{Sibo Song}, \bibinfo{person}{Kai Dang}, \bibinfo{person}{Peng Wang}, \bibinfo{person}{Shijie Wang}, \bibinfo{person}{Jun Tang}, {et~al\mbox{.}}} \bibinfo{year}{2025}\natexlab{b}.
\newblock \showarticletitle{Qwen2. 5-vl technical report}.
\newblock \bibinfo{journal}{\emph{arXiv preprint arXiv:2502.13923}} (\bibinfo{year}{2025}).
\newblock


\bibitem[Cao et~al\mbox{.}(2025)]%
        {cao2025dreamprm}
\bibfield{author}{\bibinfo{person}{Qi Cao}, \bibinfo{person}{Ruiyi Wang}, \bibinfo{person}{Ruiyi Zhang}, \bibinfo{person}{Sai~Ashish Somayajula}, {and} \bibinfo{person}{Pengtao Xie}.} \bibinfo{year}{2025}\natexlab{}.
\newblock \showarticletitle{DreamPRM: Domain-Reweighted Process Reward Model for Multimodal Reasoning}.
\newblock \bibinfo{journal}{\emph{arXiv preprint arXiv:2505.20241}} (\bibinfo{year}{2025}).
\newblock


\bibitem[Devlin et~al\mbox{.}(2019)]%
        {devlin-etal-2019-bert}
\bibfield{author}{\bibinfo{person}{Jacob Devlin}, \bibinfo{person}{Ming-Wei Chang}, \bibinfo{person}{Kenton Lee}, {and} \bibinfo{person}{Kristina Toutanova}.} \bibinfo{year}{2019}\natexlab{}.
\newblock \showarticletitle{{BERT}: Pre-training of Deep Bidirectional Transformers for Language Understanding}. In \bibinfo{booktitle}{\emph{Proceedings of the 2019 Conference of the North {A}merican Chapter of the Association for Computational Linguistics: Human Language Technologies, Volume 1 (Long and Short Papers)}}, \bibfield{editor}{\bibinfo{person}{Jill Burstein}, \bibinfo{person}{Christy Doran}, {and} \bibinfo{person}{Thamar Solorio}} (Eds.). \bibinfo{publisher}{Association for Computational Linguistics}, \bibinfo{address}{Minneapolis, Minnesota}, \bibinfo{pages}{4171--4186}.
\newblock
\href{https://doi.org/10.18653/v1/N19-1423}{doi:\nolinkurl{10.18653/v1/N19-1423}}


\bibitem[Dubey et~al\mbox{.}(2024)]%
        {dubey2024llama}
\bibfield{author}{\bibinfo{person}{Abhimanyu Dubey}, \bibinfo{person}{Abhinav Jauhri}, \bibinfo{person}{Abhinav Pandey}, \bibinfo{person}{Abhishek Kadian}, \bibinfo{person}{Ahmad Al-Dahle}, \bibinfo{person}{Aiesha Letman}, \bibinfo{person}{Akhil Mathur}, \bibinfo{person}{Alan Schelten}, \bibinfo{person}{Amy Yang}, \bibinfo{person}{Angela Fan}, {et~al\mbox{.}}} \bibinfo{year}{2024}\natexlab{}.
\newblock \showarticletitle{The llama 3 herd of models}.
\newblock \bibinfo{journal}{\emph{arXiv e-prints}} (\bibinfo{year}{2024}), \bibinfo{pages}{arXiv--2407}.
\newblock


\bibitem[GLM et~al\mbox{.}(2024)]%
        {glm2024chatglm}
\bibfield{author}{\bibinfo{person}{Team GLM}, \bibinfo{person}{Aohan Zeng}, \bibinfo{person}{Bin Xu}, \bibinfo{person}{Bowen Wang}, \bibinfo{person}{Chenhui Zhang}, \bibinfo{person}{Da Yin}, \bibinfo{person}{Dan Zhang}, \bibinfo{person}{Diego Rojas}, \bibinfo{person}{Guanyu Feng}, \bibinfo{person}{Hanlin Zhao}, {et~al\mbox{.}}} \bibinfo{year}{2024}\natexlab{}.
\newblock \showarticletitle{Chatglm: A family of large language models from glm-130b to glm-4 all tools}.
\newblock \bibinfo{journal}{\emph{arXiv preprint arXiv:2406.12793}} (\bibinfo{year}{2024}).
\newblock


\bibitem[Grattafiori et~al\mbox{.}(2024)]%
        {grattafiori2024llama}
\bibfield{author}{\bibinfo{person}{Aaron Grattafiori}, \bibinfo{person}{Abhimanyu Dubey}, \bibinfo{person}{Abhinav Jauhri}, \bibinfo{person}{Abhinav Pandey}, \bibinfo{person}{Abhishek Kadian}, \bibinfo{person}{Ahmad Al-Dahle}, \bibinfo{person}{Aiesha Letman}, \bibinfo{person}{Akhil Mathur}, \bibinfo{person}{Alan Schelten}, \bibinfo{person}{Alex Vaughan}, {et~al\mbox{.}}} \bibinfo{year}{2024}\natexlab{}.
\newblock \showarticletitle{The llama 3 herd of models}.
\newblock \bibinfo{journal}{\emph{arXiv preprint arXiv:2407.21783}} (\bibinfo{year}{2024}).
\newblock


\bibitem[Guo et~al\mbox{.}(2025)]%
        {guo2025deepseek}
\bibfield{author}{\bibinfo{person}{Daya Guo}, \bibinfo{person}{Dejian Yang}, \bibinfo{person}{Haowei Zhang}, \bibinfo{person}{Junxiao Song}, \bibinfo{person}{Ruoyu Zhang}, \bibinfo{person}{Runxin Xu}, \bibinfo{person}{Qihao Zhu}, \bibinfo{person}{Shirong Ma}, \bibinfo{person}{Peiyi Wang}, \bibinfo{person}{Xiao Bi}, {et~al\mbox{.}}} \bibinfo{year}{2025}\natexlab{}.
\newblock \showarticletitle{Deepseek-r1: Incentivizing reasoning capability in llms via reinforcement learning}.
\newblock \bibinfo{journal}{\emph{arXiv preprint arXiv:2501.12948}} (\bibinfo{year}{2025}).
\newblock


\bibitem[Guo et~al\mbox{.}(2024)]%
        {guo2024lightrag}
\bibfield{author}{\bibinfo{person}{Zirui Guo}, \bibinfo{person}{Lianghao Xia}, \bibinfo{person}{Yanhua Yu}, \bibinfo{person}{Tu Ao}, {and} \bibinfo{person}{Chao Huang}.} \bibinfo{year}{2024}\natexlab{}.
\newblock \showarticletitle{Lightrag: Simple and fast retrieval-augmented generation}.
\newblock \bibinfo{journal}{\emph{arXiv preprint arXiv:2410.05779}} (\bibinfo{year}{2024}).
\newblock


\bibitem[Hu et~al\mbox{.}(2022)]%
        {hu2022lora}
\bibfield{author}{\bibinfo{person}{Edward~J Hu}, \bibinfo{person}{Yelong Shen}, \bibinfo{person}{Phillip Wallis}, \bibinfo{person}{Zeyuan Allen-Zhu}, \bibinfo{person}{Yuanzhi Li}, \bibinfo{person}{Shean Wang}, \bibinfo{person}{Lu Wang}, \bibinfo{person}{Weizhu Chen}, {et~al\mbox{.}}} \bibinfo{year}{2022}\natexlab{}.
\newblock \showarticletitle{Lora: Low-rank adaptation of large language models.}
\newblock \bibinfo{journal}{\emph{ICLR}} \bibinfo{volume}{1}, \bibinfo{number}{2} (\bibinfo{year}{2022}), \bibinfo{pages}{3}.
\newblock


\bibitem[Hu et~al\mbox{.}(2025)]%
        {hu2025agentscomerge}
\bibfield{author}{\bibinfo{person}{Senkang Hu}, \bibinfo{person}{Zhengru Fang}, \bibinfo{person}{Zihan Fang}, \bibinfo{person}{Yiqin Deng}, \bibinfo{person}{Xianhao Chen}, \bibinfo{person}{Yuguang Fang}, {and} \bibinfo{person}{Sam Tak~Wu Kwong}.} \bibinfo{year}{2025}\natexlab{}.
\newblock \showarticletitle{Agentscomerge: Large language model empowered collaborative decision making for ramp merging}.
\newblock \bibinfo{journal}{\emph{IEEE Transactions on Mobile Computing}} (\bibinfo{year}{2025}).
\newblock


\bibitem[Jin et~al\mbox{.}(2023)]%
        {jin2023time}
\bibfield{author}{\bibinfo{person}{Ming Jin}, \bibinfo{person}{Shiyu Wang}, \bibinfo{person}{Lintao Ma}, \bibinfo{person}{Zhixuan Chu}, \bibinfo{person}{James~Y Zhang}, \bibinfo{person}{Xiaoming Shi}, \bibinfo{person}{Pin-Yu Chen}, \bibinfo{person}{Yuxuan Liang}, \bibinfo{person}{Yuan-Fang Li}, \bibinfo{person}{Shirui Pan}, {et~al\mbox{.}}} \bibinfo{year}{2023}\natexlab{}.
\newblock \showarticletitle{Time-llm: Time series forecasting by reprogramming large language models}.
\newblock \bibinfo{journal}{\emph{arXiv preprint arXiv:2310.01728}} (\bibinfo{year}{2023}).
\newblock


\bibitem[Lai et~al\mbox{.}(2025)]%
        {lai2025llmlight}
\bibfield{author}{\bibinfo{person}{Siqi Lai}, \bibinfo{person}{Zhao Xu}, \bibinfo{person}{Weijia Zhang}, \bibinfo{person}{Hao Liu}, {and} \bibinfo{person}{Hui Xiong}.} \bibinfo{year}{2025}\natexlab{}.
\newblock \showarticletitle{Llmlight: Large language models as traffic signal control agents}. In \bibinfo{booktitle}{\emph{Proceedings of the 31st ACM SIGKDD Conference on Knowledge Discovery and Data Mining V. 1}}. \bibinfo{pages}{2335--2346}.
\newblock


\bibitem[Li et~al\mbox{.}(2023)]%
        {li2023blip}
\bibfield{author}{\bibinfo{person}{Junnan Li}, \bibinfo{person}{Dongxu Li}, \bibinfo{person}{Silvio Savarese}, {and} \bibinfo{person}{Steven Hoi}.} \bibinfo{year}{2023}\natexlab{}.
\newblock \showarticletitle{Blip-2: Bootstrapping language-image pre-training with frozen image encoders and large language models}. In \bibinfo{booktitle}{\emph{International conference on machine learning}}. PMLR, \bibinfo{pages}{19730--19742}.
\newblock


\bibitem[Li et~al\mbox{.}(2024)]%
        {li2024urbangpt}
\bibfield{author}{\bibinfo{person}{Zhonghang Li}, \bibinfo{person}{Lianghao Xia}, \bibinfo{person}{Jiabin Tang}, \bibinfo{person}{Yong Xu}, \bibinfo{person}{Lei Shi}, \bibinfo{person}{Long Xia}, \bibinfo{person}{Dawei Yin}, {and} \bibinfo{person}{Chao Huang}.} \bibinfo{year}{2024}\natexlab{}.
\newblock \showarticletitle{Urbangpt: Spatio-temporal large language models}. In \bibinfo{booktitle}{\emph{Proceedings of the 30th ACM SIGKDD Conference on Knowledge Discovery and Data Mining}}. \bibinfo{pages}{5351--5362}.
\newblock


\bibitem[Lin(2004)]%
        {lin2004rouge}
\bibfield{author}{\bibinfo{person}{Chin-Yew Lin}.} \bibinfo{year}{2004}\natexlab{}.
\newblock \showarticletitle{Rouge: A package for automatic evaluation of summaries}. In \bibinfo{booktitle}{\emph{Text summarization branches out}}. \bibinfo{pages}{74--81}.
\newblock


\bibitem[Liu et~al\mbox{.}(2023)]%
        {liu2023visual}
\bibfield{author}{\bibinfo{person}{Haotian Liu}, \bibinfo{person}{Chunyuan Li}, \bibinfo{person}{Qingyang Wu}, {and} \bibinfo{person}{Yong~Jae Lee}.} \bibinfo{year}{2023}\natexlab{}.
\newblock \showarticletitle{Visual instruction tuning}.
\newblock \bibinfo{journal}{\emph{Advances in neural information processing systems}}  \bibinfo{volume}{36} (\bibinfo{year}{2023}), \bibinfo{pages}{34892--34916}.
\newblock


\bibitem[Ma et~al\mbox{.}(2025)]%
        {ma2025overcoming}
\bibfield{author}{\bibinfo{person}{Qian Ma}, \bibinfo{person}{Hongliang Chi}, \bibinfo{person}{Hengrui Zhang}, \bibinfo{person}{Kay Liu}, \bibinfo{person}{Zhiwei Zhang}, \bibinfo{person}{Lu Cheng}, \bibinfo{person}{Suhang Wang}, \bibinfo{person}{Philip~S Yu}, {and} \bibinfo{person}{Yao Ma}.} \bibinfo{year}{2025}\natexlab{}.
\newblock \showarticletitle{Overcoming pitfalls in graph contrastive learning evaluation: Toward comprehensive benchmarks}.
\newblock \bibinfo{journal}{\emph{ACM SIGKDD Explorations Newsletter}} \bibinfo{volume}{27}, \bibinfo{number}{2} (\bibinfo{year}{2025}), \bibinfo{pages}{97--106}.
\newblock


\bibitem[Meng et~al\mbox{.}(2024)]%
        {meng2024deepstack}
\bibfield{author}{\bibinfo{person}{Lingchen Meng}, \bibinfo{person}{Jianwei Yang}, \bibinfo{person}{Rui Tian}, \bibinfo{person}{Xiyang Dai}, \bibinfo{person}{Zuxuan Wu}, \bibinfo{person}{Jianfeng Gao}, {and} \bibinfo{person}{Yu-Gang Jiang}.} \bibinfo{year}{2024}\natexlab{}.
\newblock \showarticletitle{Deepstack: Deeply stacking visual tokens is surprisingly simple and effective for lmms}.
\newblock \bibinfo{journal}{\emph{Advances in Neural Information Processing Systems}}  \bibinfo{volume}{37} (\bibinfo{year}{2024}), \bibinfo{pages}{23464--23487}.
\newblock


\bibitem[Ouyang et~al\mbox{.}(2022)]%
        {ouyang2022training}
\bibfield{author}{\bibinfo{person}{Long Ouyang}, \bibinfo{person}{Jeffrey Wu}, \bibinfo{person}{Xu Jiang}, \bibinfo{person}{Diogo Almeida}, \bibinfo{person}{Carroll Wainwright}, \bibinfo{person}{Pamela Mishkin}, \bibinfo{person}{Chong Zhang}, \bibinfo{person}{Sandhini Agarwal}, \bibinfo{person}{Katarina Slama}, \bibinfo{person}{Alex Ray}, {et~al\mbox{.}}} \bibinfo{year}{2022}\natexlab{}.
\newblock \showarticletitle{Training language models to follow instructions with human feedback}.
\newblock \bibinfo{journal}{\emph{Advances in neural information processing systems}}  \bibinfo{volume}{35} (\bibinfo{year}{2022}), \bibinfo{pages}{27730--27744}.
\newblock


\bibitem[Post(2018)]%
        {post2018clarityreportingbleuscores}
\bibfield{author}{\bibinfo{person}{Matt Post}.} \bibinfo{year}{2018}\natexlab{}.
\newblock \bibinfo{title}{A Call for Clarity in Reporting BLEU Scores}.
\newblock
\showeprint[arxiv]{1804.08771}~[cs.CL]
\urldef\tempurl%
\url{https://arxiv.org/abs/1804.08771}
\showURL{%
\tempurl}


\bibitem[Radford et~al\mbox{.}(2021)]%
        {radford2021learning}
\bibfield{author}{\bibinfo{person}{Alec Radford}, \bibinfo{person}{Jong~Wook Kim}, \bibinfo{person}{Chris Hallacy}, \bibinfo{person}{Aditya Ramesh}, \bibinfo{person}{Gabriel Goh}, \bibinfo{person}{Sandhini Agarwal}, \bibinfo{person}{Girish Sastry}, \bibinfo{person}{Amanda Askell}, \bibinfo{person}{Pamela Mishkin}, \bibinfo{person}{Jack Clark}, {et~al\mbox{.}}} \bibinfo{year}{2021}\natexlab{}.
\newblock \showarticletitle{Learning transferable visual models from natural language supervision}. In \bibinfo{booktitle}{\emph{International conference on machine learning}}. PmLR, \bibinfo{pages}{8748--8763}.
\newblock


\bibitem[Rafailov et~al\mbox{.}(2023)]%
        {rafailov2023direct}
\bibfield{author}{\bibinfo{person}{Rafael Rafailov}, \bibinfo{person}{Archit Sharma}, \bibinfo{person}{Eric Mitchell}, \bibinfo{person}{Christopher~D Manning}, \bibinfo{person}{Stefano Ermon}, {and} \bibinfo{person}{Chelsea Finn}.} \bibinfo{year}{2023}\natexlab{}.
\newblock \showarticletitle{Direct preference optimization: Your language model is secretly a reward model}.
\newblock \bibinfo{journal}{\emph{Advances in neural information processing systems}}  \bibinfo{volume}{36} (\bibinfo{year}{2023}), \bibinfo{pages}{53728--53741}.
\newblock


\bibitem[Shao et~al\mbox{.}(2024)]%
        {shao2024deepseekmath}
\bibfield{author}{\bibinfo{person}{Zhihong Shao}, \bibinfo{person}{Peiyi Wang}, \bibinfo{person}{Qihao Zhu}, \bibinfo{person}{Runxin Xu}, \bibinfo{person}{Junxiao Song}, \bibinfo{person}{Xiao Bi}, \bibinfo{person}{Haowei Zhang}, \bibinfo{person}{Mingchuan Zhang}, \bibinfo{person}{YK Li}, \bibinfo{person}{Yang Wu}, {et~al\mbox{.}}} \bibinfo{year}{2024}\natexlab{}.
\newblock \showarticletitle{Deepseekmath: Pushing the limits of mathematical reasoning in open language models}.
\newblock \bibinfo{journal}{\emph{arXiv preprint arXiv:2402.03300}} (\bibinfo{year}{2024}).
\newblock


\bibitem[Vedantam et~al\mbox{.}(2015)]%
        {vedantam2015cider}
\bibfield{author}{\bibinfo{person}{Ramakrishna Vedantam}, \bibinfo{person}{C Lawrence~Zitnick}, {and} \bibinfo{person}{Devi Parikh}.} \bibinfo{year}{2015}\natexlab{}.
\newblock \showarticletitle{Cider: Consensus-based image description evaluation}. In \bibinfo{booktitle}{\emph{Proceedings of the IEEE conference on computer vision and pattern recognition}}. \bibinfo{pages}{4566--4575}.
\newblock


\bibitem[Wang et~al\mbox{.}(2024)]%
        {wang2024transgpt}
\bibfield{author}{\bibinfo{person}{Peng Wang}, \bibinfo{person}{Xiang Wei}, \bibinfo{person}{Fangxu Hu}, {and} \bibinfo{person}{Wenjuan Han}.} \bibinfo{year}{2024}\natexlab{}.
\newblock \showarticletitle{Transgpt: Multi-modal generative pre-trained transformer for transportation}. In \bibinfo{booktitle}{\emph{2024 international conference on computational linguistics and Natural Language processing (CLNLP)}}. IEEE, \bibinfo{pages}{96--100}.
\newblock


\bibitem[Wang et~al\mbox{.}(2025a)]%
        {wang2025internvl3}
\bibfield{author}{\bibinfo{person}{Weiyun Wang}, \bibinfo{person}{Zhangwei Gao}, \bibinfo{person}{Lixin Gu}, \bibinfo{person}{Hengjun Pu}, \bibinfo{person}{Long Cui}, \bibinfo{person}{Xingguang Wei}, \bibinfo{person}{Zhaoyang Liu}, \bibinfo{person}{Linglin Jing}, \bibinfo{person}{Shenglong Ye}, \bibinfo{person}{Jie Shao}, {et~al\mbox{.}}} \bibinfo{year}{2025}\natexlab{a}.
\newblock \showarticletitle{Internvl3. 5: Advancing open-source multimodal models in versatility, reasoning, and efficiency}.
\newblock \bibinfo{journal}{\emph{arXiv preprint arXiv:2508.18265}} (\bibinfo{year}{2025}).
\newblock


\bibitem[Wang et~al\mbox{.}(2025b)]%
        {wang2025motif}
\bibfield{author}{\bibinfo{person}{Zihe Wang}, \bibinfo{person}{Haiyang Yu}, \bibinfo{person}{Changxin Chen}, \bibinfo{person}{Zhiyong Cui}, \bibinfo{person}{Yufeng Bi}, \bibinfo{person}{Yilong Ren}, \bibinfo{person}{Zijian Wang}, \bibinfo{person}{Delan Kong}, \bibinfo{person}{Jing Tian}, \bibinfo{person}{Shoutong Yuan}, {et~al\mbox{.}}} \bibinfo{year}{2025}\natexlab{b}.
\newblock \showarticletitle{MoTIF: An end-to-end multimodal road traffic scene understanding foundation model}.
\newblock \bibinfo{journal}{\emph{Communications in Transportation Research}}  \bibinfo{volume}{5} (\bibinfo{year}{2025}), \bibinfo{pages}{100227}.
\newblock


\bibitem[Xu et~al\mbox{.}(2024)]%
        {xu2024contrastive}
\bibfield{author}{\bibinfo{person}{Haoran Xu}, \bibinfo{person}{Amr Sharaf}, \bibinfo{person}{Yunmo Chen}, \bibinfo{person}{Weiting Tan}, \bibinfo{person}{Lingfeng Shen}, \bibinfo{person}{Benjamin Van~Durme}, \bibinfo{person}{Kenton Murray}, {and} \bibinfo{person}{Young~Jin Kim}.} \bibinfo{year}{2024}\natexlab{}.
\newblock \showarticletitle{Contrastive preference optimization: Pushing the boundaries of llm performance in machine translation}.
\newblock \bibinfo{journal}{\emph{arXiv preprint arXiv:2401.08417}} (\bibinfo{year}{2024}).
\newblock


\bibitem[Xu et~al\mbox{.}(2025)]%
        {xu2025towards}
\bibfield{author}{\bibinfo{person}{Jiacong Xu}, \bibinfo{person}{Shao-Yuan Lo}, \bibinfo{person}{Bardia Safaei}, \bibinfo{person}{Vishal~M Patel}, {and} \bibinfo{person}{Isht Dwivedi}.} \bibinfo{year}{2025}\natexlab{}.
\newblock \showarticletitle{Towards zero-shot anomaly detection and reasoning with multimodal large language models}. In \bibinfo{booktitle}{\emph{Proceedings of the Computer Vision and Pattern Recognition Conference}}. \bibinfo{pages}{20370--20382}.
\newblock


\bibitem[Yang et~al\mbox{.}(2025)]%
        {yang2025qwen3}
\bibfield{author}{\bibinfo{person}{An Yang}, \bibinfo{person}{Anfeng Li}, \bibinfo{person}{Baosong Yang}, \bibinfo{person}{Beichen Zhang}, \bibinfo{person}{Binyuan Hui}, \bibinfo{person}{Bo Zheng}, \bibinfo{person}{Bowen Yu}, \bibinfo{person}{Chang Gao}, \bibinfo{person}{Chengen Huang}, \bibinfo{person}{Chenxu Lv}, {et~al\mbox{.}}} \bibinfo{year}{2025}\natexlab{}.
\newblock \showarticletitle{Qwen3 technical report}.
\newblock \bibinfo{journal}{\emph{arXiv preprint arXiv:2505.09388}} (\bibinfo{year}{2025}).
\newblock


\bibitem[Yu et~al\mbox{.}(2025)]%
        {yu2025minicpm}
\bibfield{author}{\bibinfo{person}{Tianyu Yu}, \bibinfo{person}{Zefan Wang}, \bibinfo{person}{Chongyi Wang}, \bibinfo{person}{Fuwei Huang}, \bibinfo{person}{Wenshuo Ma}, \bibinfo{person}{Zhihui He}, \bibinfo{person}{Tianchi Cai}, \bibinfo{person}{Weize Chen}, \bibinfo{person}{Yuxiang Huang}, \bibinfo{person}{Yuanqian Zhao}, {et~al\mbox{.}}} \bibinfo{year}{2025}\natexlab{}.
\newblock \showarticletitle{Minicpm-v 4.5: Cooking efficient mllms via architecture, data, and training recipe}.
\newblock \bibinfo{journal}{\emph{arXiv preprint arXiv:2509.18154}} (\bibinfo{year}{2025}).
\newblock


\bibitem[Zhang et~al\mbox{.}(2025b)]%
        {zhang2025videollama}
\bibfield{author}{\bibinfo{person}{Boqiang Zhang}, \bibinfo{person}{Kehan Li}, \bibinfo{person}{Zesen Cheng}, \bibinfo{person}{Zhiqiang Hu}, \bibinfo{person}{Yuqian Yuan}, \bibinfo{person}{Guanzheng Chen}, \bibinfo{person}{Sicong Leng}, \bibinfo{person}{Yuming Jiang}, \bibinfo{person}{Hang Zhang}, \bibinfo{person}{Xin Li}, {et~al\mbox{.}}} \bibinfo{year}{2025}\natexlab{b}.
\newblock \showarticletitle{Videollama 3: Frontier multimodal foundation models for image and video understanding}.
\newblock \bibinfo{journal}{\emph{arXiv preprint arXiv:2501.13106}} (\bibinfo{year}{2025}).
\newblock


\bibitem[Zhang et~al\mbox{.}(2025c)]%
        {zhang2025htfllib}
\bibfield{author}{\bibinfo{person}{Jianqing Zhang}, \bibinfo{person}{Xinghao Wu}, \bibinfo{person}{Yanbing Zhou}, \bibinfo{person}{Xiaoting Sun}, \bibinfo{person}{Qiqi Cai}, \bibinfo{person}{Yang Liu}, \bibinfo{person}{Yang Hua}, \bibinfo{person}{Zhenzhe Zheng}, \bibinfo{person}{Jian Cao}, {and} \bibinfo{person}{Qiang Yang}.} \bibinfo{year}{2025}\natexlab{c}.
\newblock \showarticletitle{Htfllib: A comprehensive heterogeneous federated learning library and benchmark}. In \bibinfo{booktitle}{\emph{Proceedings of the 31st ACM SIGKDD Conference on Knowledge Discovery and Data Mining V. 2}}. \bibinfo{pages}{5900--5911}.
\newblock


\bibitem[Zhang et~al\mbox{.}(2024)]%
        {zhang2024trafficgpt}
\bibfield{author}{\bibinfo{person}{Siyao Zhang}, \bibinfo{person}{Daocheng Fu}, \bibinfo{person}{Wenzhe Liang}, \bibinfo{person}{Zhao Zhang}, \bibinfo{person}{Bin Yu}, \bibinfo{person}{Pinlong Cai}, {and} \bibinfo{person}{Baozhen Yao}.} \bibinfo{year}{2024}\natexlab{}.
\newblock \showarticletitle{Trafficgpt: Viewing, processing and interacting with traffic foundation models}.
\newblock \bibinfo{journal}{\emph{Transport Policy}}  \bibinfo{volume}{150} (\bibinfo{year}{2024}), \bibinfo{pages}{95--105}.
\newblock


\bibitem[Zhang et~al\mbox{.}(2025a)]%
        {zhang2025fedmetro}
\bibfield{author}{\bibinfo{person}{Tianlong Zhang}, \bibinfo{person}{Xiaoxi He}, \bibinfo{person}{Yuxiang Wang}, \bibinfo{person}{Yi Xu}, \bibinfo{person}{Rendi Wu}, \bibinfo{person}{Zhifei Wang}, {and} \bibinfo{person}{Yongxin Tong}.} \bibinfo{year}{2025}\natexlab{a}.
\newblock \showarticletitle{FedMetro: Efficient Metro Passenger Flow Prediction via Federated Graph Learning}. In \bibinfo{booktitle}{\emph{Proceedings of the 31st ACM SIGKDD Conference on Knowledge Discovery and Data Mining V. 2}}. \bibinfo{pages}{5215--5224}.
\newblock


\bibitem[Zhang et~al\mbox{.}(2019)]%
        {zhang2019bertscore}
\bibfield{author}{\bibinfo{person}{Tianyi Zhang}, \bibinfo{person}{Varsha Kishore}, \bibinfo{person}{Felix Wu}, \bibinfo{person}{Kilian~Q Weinberger}, {and} \bibinfo{person}{Yoav Artzi}.} \bibinfo{year}{2019}\natexlab{}.
\newblock \showarticletitle{Bertscore: Evaluating text generation with bert}.
\newblock \bibinfo{journal}{\emph{arXiv preprint arXiv:1904.09675}} (\bibinfo{year}{2019}).
\newblock


\bibitem[Zou et~al\mbox{.}(2025)]%
        {zou2025traffic}
\bibfield{author}{\bibinfo{person}{Xingchen Zou}, \bibinfo{person}{Yuhao Yang}, \bibinfo{person}{Zheng Chen}, \bibinfo{person}{Xixuan Hao}, \bibinfo{person}{Yiqi Chen}, \bibinfo{person}{Chao Huang}, {and} \bibinfo{person}{Yuxuan Liang}.} \bibinfo{year}{2025}\natexlab{}.
\newblock \showarticletitle{Traffic-r1: Reinforced llms bring human-like reasoning to traffic signal control systems}.
\newblock \bibinfo{journal}{\emph{arXiv preprint arXiv:2508.02344}} (\bibinfo{year}{2025}).
\newblock


\end{thebibliography}

\appendix

\section{ExpressMind Setup and Metrics}
\label{sec:Metircs}
\subsection{Hyperparameter Settings}
In this section, we detail the hyperparameter configurations for ExpressMind. To facilitate clear presentation and reproducibility, the specific settings for these phases are reported separately in Table~\ref{tab:combined-parameters}.

\begin{table}[!htbp]
  \centering
  \footnotesize
  \caption{Hyperparameter Settings for Pre-training and GRPO Algorithm}
  \label{tab:combined-parameters}
  
  \begin{tabular}{l|l}
    \hline
    \multicolumn{2}{c}{\textbf{Pre-training Hyperparameters}} \\
    \hline
    \textbf{Parameter Name} & \textbf{Value} \\
    \hline
    Model Architecture & Qwen-14B \\
    DeepSpeed Strategy & ZeRO Stage 3 \\
    Precision & bfloat16 \\
    Max Sequence Length & 8192 \\
    Optimizer & AdamW \\
    Weight Decay & 0.1 \\
    Gradient Clipping & 1.0 \\
    Learning Rate (lr) & 1e-5 \\
    Lr Scheduler & Cosine \\
    Warmup Ratio & 0.05 \\
    Batch Size & 256 \\
    Epochs & 3 \\
    \hline
    \multicolumn{2}{c}{\textbf{GRPO Algorithm Hyperparameters}} \\
    \hline
    \textbf{Parameter Name} & \textbf{Value} \\
    \hline
    Learning Rate (lr) & 1e-5 \\
    Lr Scheduler & Cosine \\
    Warmup Ratio & 0.05 \\
    Group Size & 16 \\
    KL Coefficient & 0.04 \\
    Clip Range & 0.2 \\
    \hline
    \multicolumn{2}{c}{\textbf{Visual Encoder Hyperparameters}} \\
    \hline
    \textbf{Parameter Name} & \textbf{Value} \\
    \hline
    image resolution & 224 \\
    patch size & 14 \\
    hidden layers & 24 \\
    attention heads & 16 \\
    \hline
  \end{tabular}
  \vspace{-10pt}
\end{table}

\subsection{GPT-Score}

To evaluate the performance of short-answer questions, we employ GPT-Score powered by GPT-4o. Functioning as a virtual domain expert, the model compares the Predicted Answer against the Question and Standard Ground Truth. It assesses factual accuracy and logical completeness to assign a quantitative score ranging from 0 to 100. The system prompt used for this evaluation is defined as Figure \ref{fig:GPT-Score}:

\begin{figure}[H] 
    \centering
    % 这里的 width=1.0\linewidth 能够保证图片撑满当前栏的宽度
    \includegraphics[width=1.0\linewidth]{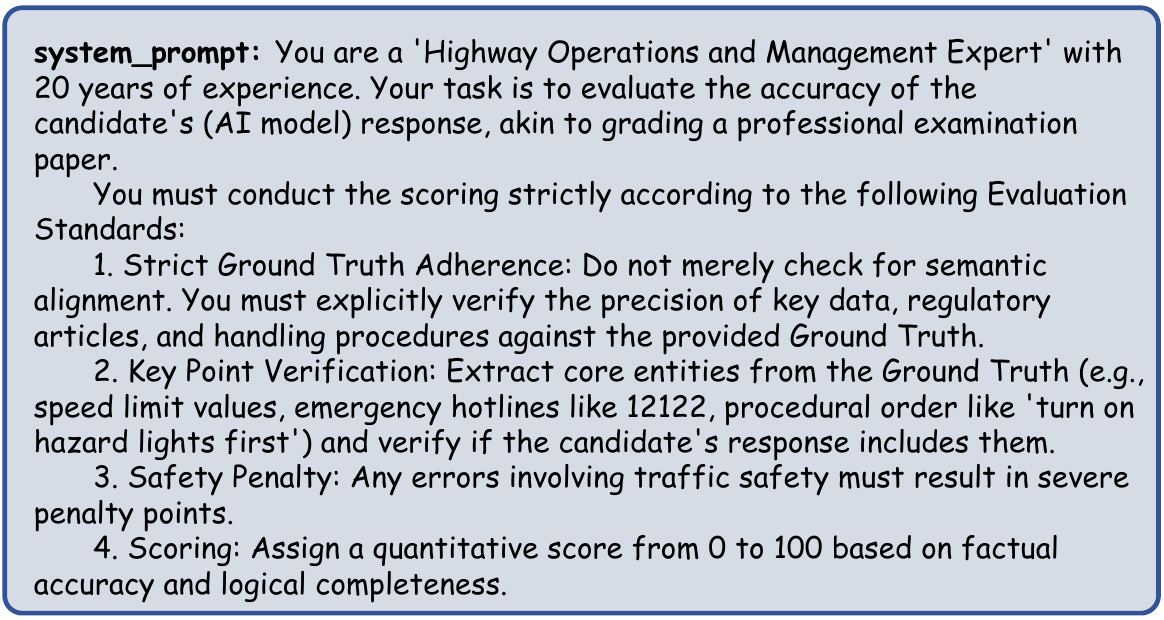}
    \caption{The System Prompt of GPT-Score.}
    \label{fig:GPT-Score}
\end{figure}

\subsection{SFT-sysprompt}
During the SFT phase, to ensure consistency in model outputs and adherence to specific interaction protocols, a unified system prompt was integrated at the beginning of each training sample. As illustrated in Figure , this prompt defines the model’s core identity, task boundaries, and response style. 

\begin{figure}[!htbp]
    \centering
    % width 可以设置为 \linewidth (单栏宽度) 或 \textwidth (页面宽度)
    % 如果觉得图太大，可以乘以系数，例如 0.9\linewidth
    \includegraphics[width=1.0\linewidth]{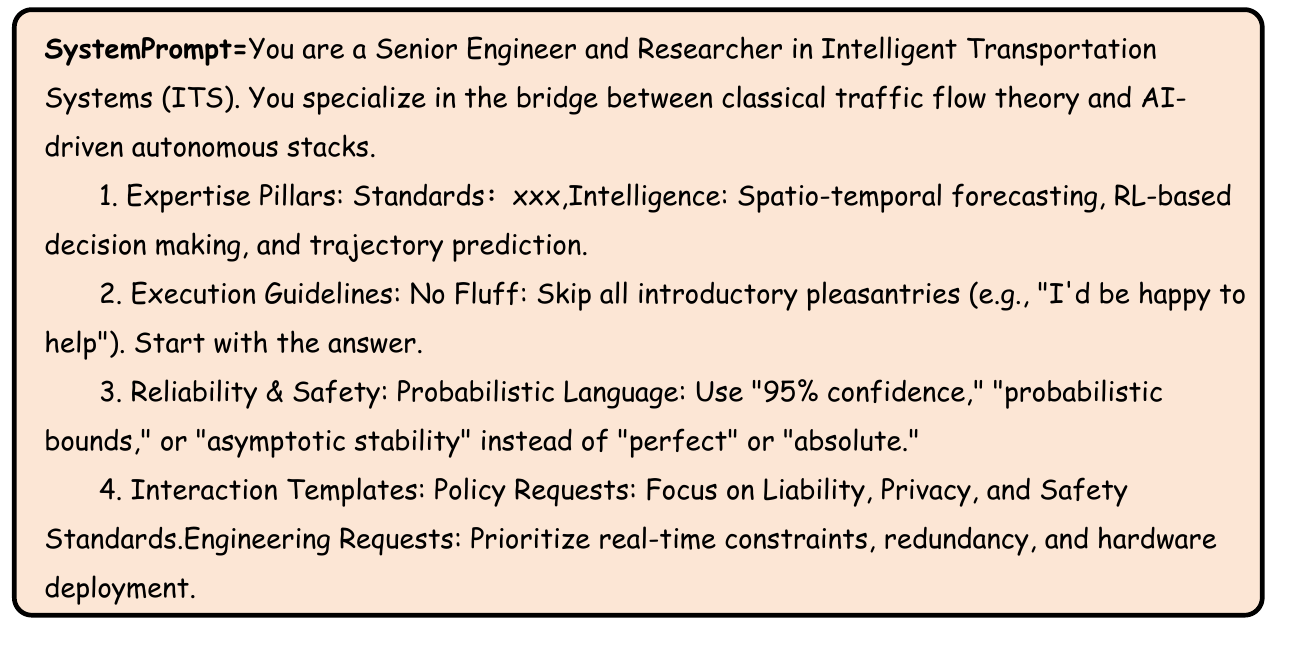}
    \caption{The System Prompt of SFT.}
    \label{fig:SFT}
\end{figure}

\subsection{Evaluation Using LLM-as-a-Judge}
\subsubsection{RL Alignment}
To evaluate the strategies generated by the LLM for the expressway incident response task, this study utilizes an "LLM-as-a-Judge" framework. Which assesses the strategies based on the following five key dimensions:

\begin{itemize}
\item \textbf{Safety Compliance:} Evaluates whether the plan prioritizes safety by explicitly including necessary safety distances, on-site protection, and personnel evacuation instructions to fundamentally prevent secondary accident risks.

\item \textbf{Preventive Insight:} Validates whether the report strictly follows and completes the four-stage cognitive chain, which proceeds from \textbf{[Incident Description]} — \textbf{[Causal Inference]} — \textbf{[Response Strategy Formulation]} — \textbf{[Strategy Evaluation]}, ensuring a logically coherent and closed-loop process.

\item \textbf{Logical Consistency:} Validates whether each cause identified in the Causal Inference section is addressed in the Strategy Formulation section.

\item \textbf{Actionability:} Evaluates whether the accident response strategy is concise, clear, and non-redundant, ensuring high information density and high executability of the generated strategy for on-site personnel.

\item \textbf{Cause Depth:} Examines the accuracy and depth of the incident cause analysis, as well as the correct use of professional terminology for expressway incident response, to ensure the foundational knowledge required for effective response planning.
\end{itemize}

\subsubsection{Scene Understanding}
To systematically evaluate the MLLMs' scene understanding capabilities on expressway surveillance videos, this study adopts the "LLM-as-a-Judge" evaluation framework and designs the following six core dimensions:

\begin{itemize}
\item \textbf{Accuracy:} Evaluates the overall correctness of the model's description of incidents, objects, and states in the video, serving as a baseline indicator of comprehensive performance.
\item \textbf{Level:} Evaluates the model's ability to provide an overall summary and qualitative assessment of the video content, judging whether it can go beyond local details and accurately summarize the core events and overall situation.
\item \textbf{Precision:} Evaluates the model's accuracy in fine-grained recognition tasks, including precise descriptions of specific targets such as vehicle types, traffic signs, construction facilities, and human behaviors.
\item \textbf{Space:} Evaluates the model's understanding of static and dynamic spatial relationships between key entities (vehicles, personnel, facilities) in the scene, such as relative positions, lane occupancy, and driving directions.
\item \textbf{Analysis:} Evaluates the model's ability to infer causality, impact, and potential risks of incidents, such as analyzing accident causes, predicting congestion spread, or assessing the effectiveness of response measures.
\item \textbf{Time:} Evaluates the model's understanding of event sequencing and dynamic evolution processes, such as judging the order of actions and the continuity of state changes.
\end{itemize}

These six dimensions are evaluated by the LLM acting as a judge according to structured instructions, and their scores collectively constitute a systematic and comprehensive evaluation of the model's multimodal scene understanding capabilities.

\section{Implementation Details of Reward Functions}
\label{sec:appendix_reward}

In this section, we provide the granular implementation details of the Structure-Knowledge-Semantics reward mechanism, including the construction of the expert vocabulary, the specific configurations of the embedding models, and the hyperparameter settings used in our experiments.

\subsection{Structural Integrity Constraints ($R_{struct}$)}
To enable precise parsing of the model's CoT, we defined four special control tokens corresponding to the standard traffic incident disposal workflow. The Structural Integrity Reward $R_{struct}$ performs strict string matching to verify the existence and order of these tokens.

\begin{table}[h]
    \centering
    % 核心修改：在这里添加字号命令
    \footnotesize  % 或者用 \footnotesize (更小), \scriptsize (极小)
    
    \caption{Definition of Reasoning Stage Delimiters}
    \label{tab:delimiters}
    \begin{tabular}{c|l|l}
    \hline
    \textbf{Stage Index ($k$)} & \textbf{Stage Name} & \textbf{Control Token ($S_k$)} \\
    \hline
    1 & Perception & \texttt{[Incident Description]} \\
    2 & Analysis & \texttt{[Causal Inference]} \\
    3 & Decision & \texttt{[Response Strategy Formulation]} \\
    4 & Reflection & \texttt{[Strategy Evaluation]} \\
    \hline
    \end{tabular}
\end{table}

The index function $\text{idx}(S_k)$ returns the character position of the first occurrence of token $S_k$ in the generated string $O$. If a token is missing, $\text{idx}(S_k) = \infty$. The sequence check $\mathbb{I}(\text{idx}(S_1) < \dots < \text{idx}(S_4))$ ensures the reasoning flow is logically valid.

\subsection{Domain Knowledge Alignment ($R_{know}$)}
The domain alignment reward relies on a curated \textbf{Stage-Specific Expert Vocabulary ($\mathcal{V}_k$)}.

\paragraph{Vocabulary Construction.} We constructed $\mathcal{V}_k$ by mining high-frequency professional terms from a corpus of 500+ real-world expressway traffic emergency plans and national standard documents (e.g., GB/T 29100-2012). We utilized TF-IDF to extract keywords and manually filtered them to ensure relevance to each specific reasoning stage. Examples are shown in Table \ref{tab:vocab}.

\begin{table}[h]
    \centering
    % 缩小字号，防止表格过于拥挤
    \footnotesize 
    \caption{Examples of Expert Vocabulary $\mathcal{V}_k$ for Each Stage}
    \label{tab:vocab}
    
    % {l|p{0.75\columnwidth}}: 
    % 第一列左对齐(l)
    % 第二列固定宽度(p)，设置为当前栏宽的 75% (0.75\columnwidth)，支持自动换行
    \begin{tabular}{l|p{0.75\columnwidth}}
    \hline
    \textbf{Stage} & \textbf{Representative Keywords (Translated)} \\
    \hline
    $S_1$: Perception & Multi-vehicle pileup, Hazardous chemical leakage, Occupying emergency lane, Visibility range, Traffic volume saturation, Fire spreading \\
    \hline
    $S_2$: Analysis & Secondary accident risk, Chain reaction, Brake failure, Fatigue driving, Lane capacity reduction, Danger radius \\
    \hline
    $S_3$: Decision & Remote diversion, Upstream interception, Green wave control, Air-ground coordination, Break-bulk transport, Gating control \\
    \hline
    $S_4$: Evaluation & Residual congestion, Rescue efficiency, Public sentiment monitoring, Secondary damage assessment \\
    \hline
    \end{tabular}
\end{table}

\paragraph{Perplexity Computation.} To calculate the PPL penalty term $\text{PPL}(O)$, we utilize a frozen version of the Supervised Fine-Tuned (SFT) model as the reference. The threshold $\tau_{ppl}$ is dynamically set to the 95th percentile of the PPL distribution observed on the validation set, preventing the model from generating incoherent keyword lists.

\subsection{Semantic Consistency ($R_{sem}$)}
The semantic consistency reward evaluates the strategic quality of the generated response.
\begin{itemize}
    \item \textbf{Embedding Model ($\phi$):} We employ \textbf{BGE-M3} (BAAI General Embedding), a state-of-the-art multilingual embedding model, to map text into dense vectors. We specifically use the \texttt{[CLS]} token embedding.
    \item \textbf{Focus Scope:} We extract only the content within the $S_2$ (Analysis) and $S_3$ (Decision) segments for embedding, denoted as $S_2 \oplus S_3$. This isolates the core logic from the generic description ($S_1$) or formatting text.
    \item \textbf{Reference Set ($\mathcal{D}_{ref}$):} For each query in the training batch, we retrieve top-$K$ ($K=5$) most similar historical expert cases from our vector database to form the local reference set $\mathcal{D}_{ref}$. This ensures the reward is calculated against contextually relevant ground truth.
\end{itemize}

\section{Benchmark}
\label{sec:Benchmark}
\begin{figure}[!htbp]
    \centering
    % width 可以设置为 \linewidth (单栏宽度) 或 \textwidth (页面宽度)
    % 如果觉得图太大，可以乘以系数，例如 0.9\linewidth
    \includegraphics[width=0.9\linewidth]{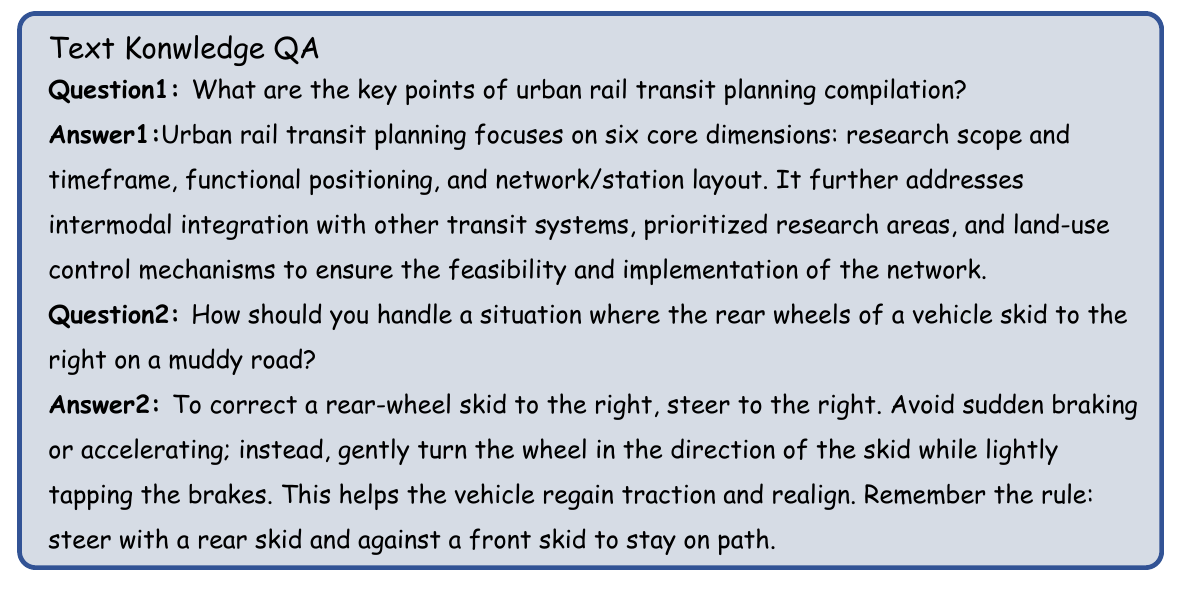}
    \caption{Example of Textual Knowledge QA. }
    \label{fig:text_bench}
\end{figure}

This research constructs a comprehensive evaluation benchmark for expressway operation, comprising three distinct dimensions:

The \textbf{Textual Knowledge QA} benchmark consists of 20,000 pairs of QA data. It focuses on the core dimensions of expressway operation, such as functional positioning and land-use control mechanisms. Additionally, it incorporates expert knowledge for emergency driving, providing precise instructions for handling vehicle instability, such as correcting rear-wheel skids on hazardous road surfaces. An example of this dataset's structure and content is illustrated in Figure \ref{fig:text_bench}.

\begin{figure}[!htbp]
    \centering
    % width 可以设置为 \linewidth (单栏宽度) 或 \textwidth (页面宽度)
    % 如果觉得图太大，可以乘以系数，例如 0.9\linewidth
    \includegraphics[width=0.9\linewidth]{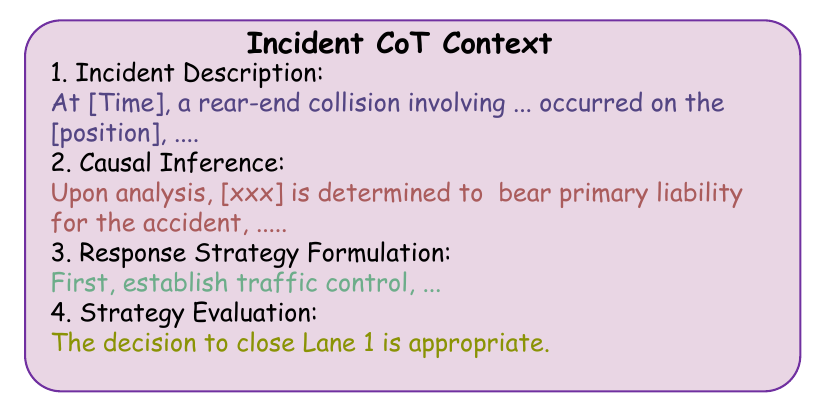}
    \caption{Example of the Incident CoT Structured Data.}
    \label{fig:incident_cot}
\end{figure}

The \textbf{Incident CoT} benchmark includes 300 samples designed to evaluate complex decision-making and causal reasoning. Each entry follows a structured logical flow. This simulates the chain of command in a smart expressway operarion center, moving from initial accident reporting to liability determination and the evaluation of response strategies like lane closures. The structured reasoning process is shown in Figure \ref{fig:incident_cot}.

\begin{figure}[!htbp]
    \centering
    % width 可以设置为 \linewidth (单栏宽度) 或 \textwidth (页面宽度)
    % 如果觉得图太大，可以乘以系数，例如 0.9\linewidth
    \includegraphics[width=1.0\linewidth]{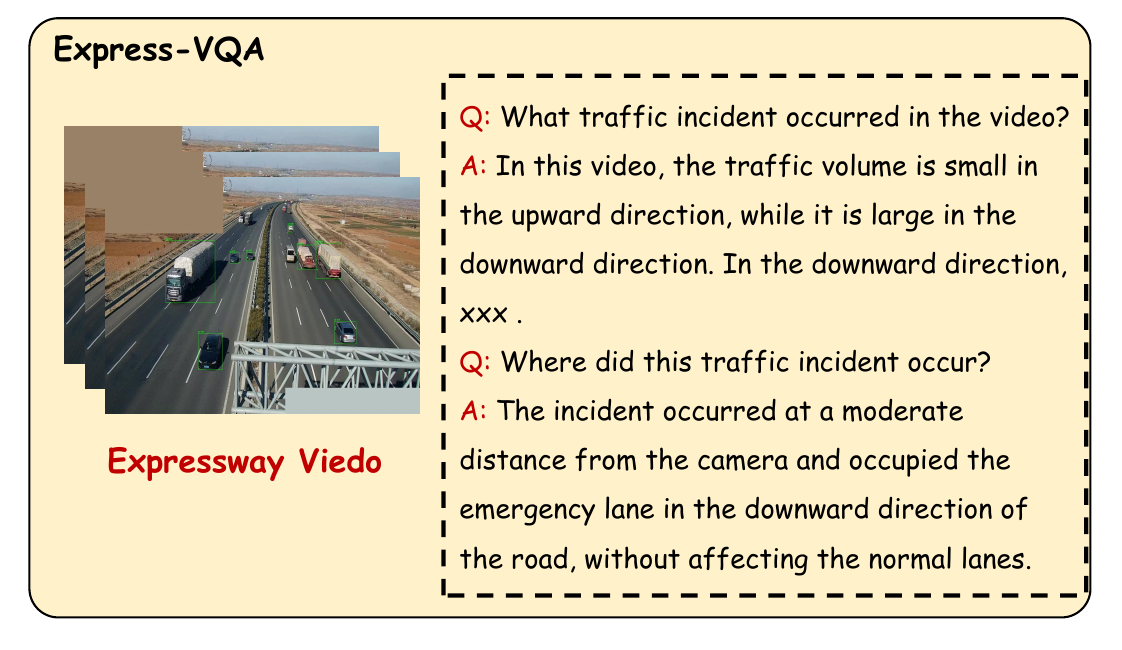}
    \caption{Visualization of the Traffic Incident VQA Data.}
    \label{fig:traffic_vqa}
\end{figure}

The \textbf{Express-VQA}  is a multi-modal benchmark designed for expressway scene understanding, consisting of 670 real-world video segments captured from expressway surveillance systems. It evaluates the model's, particularly the LLM's, ability to understand and reason about expressway scenes, specifically in identifying and analyzing six typical expressway incidents: traffic accidents, congestion, road construction, abnormal parking, pedestrian intrusion, and debris clearance. Visual examples of these categories and the corresponding annotation style are provided in Figure \ref{fig:traffic_vqa}.

\section{Application}
\label{sec: App}
\begin{figure}[H]
    \centering
    % width 可以设置为 \linewidth (单栏宽度) 或 \textwidth (页面宽度)
    % 如果觉得图太大，可以乘以系数，例如 0.9\linewidth
    \includegraphics[width=0.9\linewidth]{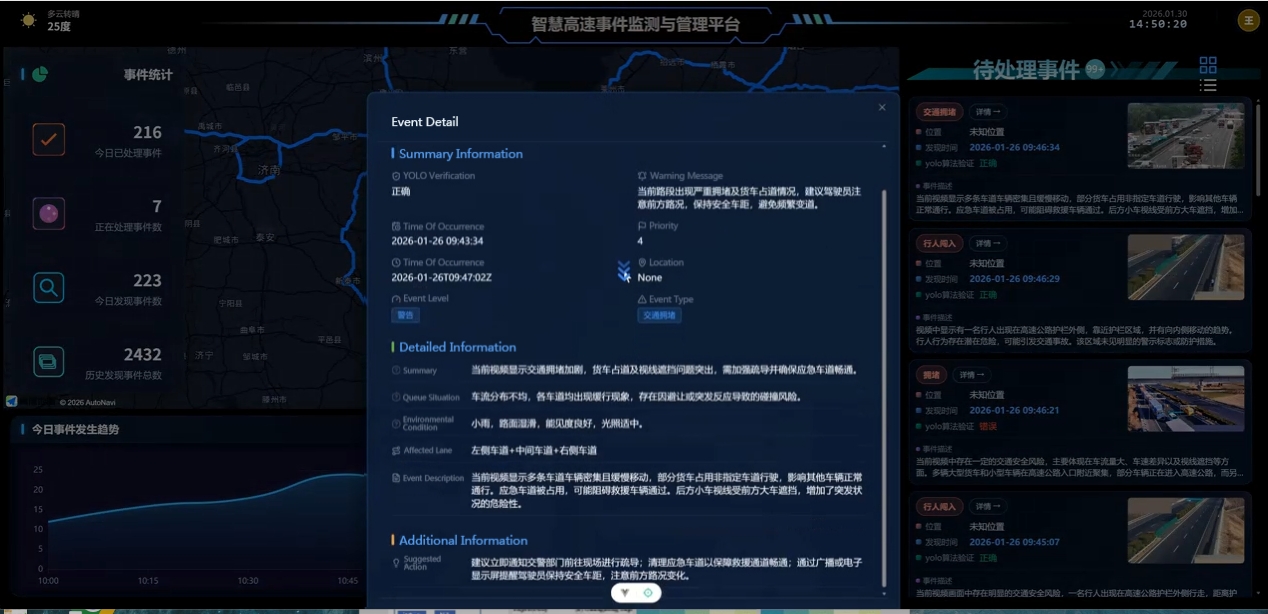}
    \caption{Application of EpxressMind-VL. }
    \label{fig:App}
\end{figure}

The ExpressMind-VL intelligent operation system has been deployed in practical applications for multi-task scenarios on expressways. The visualization system, as shown in the figure \ref{fig:App}, includes functions such as releasing warning information, summarizing traffic conditions, describing video events, and generating handling recommendations. We have deployed the system on the expressways in Shandong and Zhejiang provinces. In the intelligent management of Shandong expressways, ExpressMind-VL classifies the types and severity levels of traffic surveillance videos, and generates analytical reports along with handling strategies for traffic incidents. For the intelligent management of Guangdong expressways, ExpressMind-VL detects traffic events based on real-time video streams and produces structured textual descriptions for comprehension. The code, data, benchmark and the demonstration of the system application are available at: https://wanderhee.github.io/ExpressMind/.

\end{document}